\documentclass[10pt, twocolumn, letterpaper]{article}

\usepackage[pagenumbers]{cvpr} 

\definecolor{cvprblue}{rgb}{0.21,0.49,0.74}
\usepackage[pagebackref, breaklinks, colorlinks, allcolors=cvprblue]{hyperref}

\usepackage{amsmath}
\usepackage{amssymb}
\usepackage{amsfonts}
\usepackage{bm}
\usepackage{comment}
\usepackage{color}
\usepackage{array}
\usepackage{colortab}
\usepackage{colortbl}
\usepackage{arydshln}
\usepackage{multirow}
\usepackage{wrapfig}
\usepackage{pifont}

\usepackage[normalem]{ulem}
\useunder{\uline}{\ul}{}

\newcommand{\xmark}{\ding{55}}

\newcommand{\nbf}[1]{
\noindent
\textbf{#1}\hspace{0.5em}}
\usepackage{listings}

\def\paperID{17393} 
\def\confName{CVPR}
\def\confYear{2025}

\title{VASCAR: Content-Aware Layout Generation via Visual-Aware Self-Correction}

\author{
\textbf{Jiahao Zhang}$^{\clubsuit}$\thanks{Work done during an internship at LY Corporation.}, \textbf{Ryota Yoshihashi}$^{\diamondsuit}$, \textbf{Shunsuke Kitada}$^{\diamondsuit}$, \textbf{Atsuki Osanai}$^{\diamondsuit}$, \textbf{Yuta Nakashima}$^{\clubsuit}$ \\
$^{\clubsuit}$D3 Center, Osaka University, Japan\\
$^{\diamondsuit}$LY Corporation, Japan\\
{\tt\normalsize \{jiahao@is., n-yuta@\}ids.osaka-u.ac.jp}\\
{\tt\normalsize \{ryoshiha, s.kitada, atsuki.osanai\}@lycorp.co.jp}
}

\begin{document}
    \maketitle
    \begin{abstract}
Large language models (LLMs) have proven effective for layout generation due to their ability to produce structure-description languages, such as HTML or JSON.
In this paper, we argue that while LLMs can perform reasonably well in certain cases, their intrinsic limitation of not being able to perceive images restricts their effectiveness in tasks requiring visual content, e.g., content-aware layout generation. Therefore, we explore whether large vision-language models (LVLMs) can be applied to content-aware layout generation. To this end, inspired by the iterative revision and heuristic evaluation workflow of designers, we propose the training-free \textit{\textbf{\underline{V}}isual-\textbf{\underline{A}}ware \textbf{\underline{S}}elf-\textbf{\underline{C}}orrection L\textbf{\underline{A}}yout Gene\textbf{\underline{R}}ation} (\textbf{VASCAR}). VASCAR enables LVLMs (e.g., GPT-4o and Gemini) iteratively refine their outputs with reference to rendered layout images, which are visualized as colored bounding boxes on poster background (i.e., canvas). Extensive experiments and user study demonstrate VASCAR's effectiveness, achieving state-of-the-art (SOTA) layout generation quality. Furthermore, the generalizability of VASCAR across GPT-4o and Gemini demonstrates its versatility.
\end{abstract}
    \section{Introduction}
\label{sec:intro}

\begin{figure}[t]
  \centering
  \begin{subfigure}{0.9\linewidth}
    \centering
    \includegraphics[width=\linewidth]{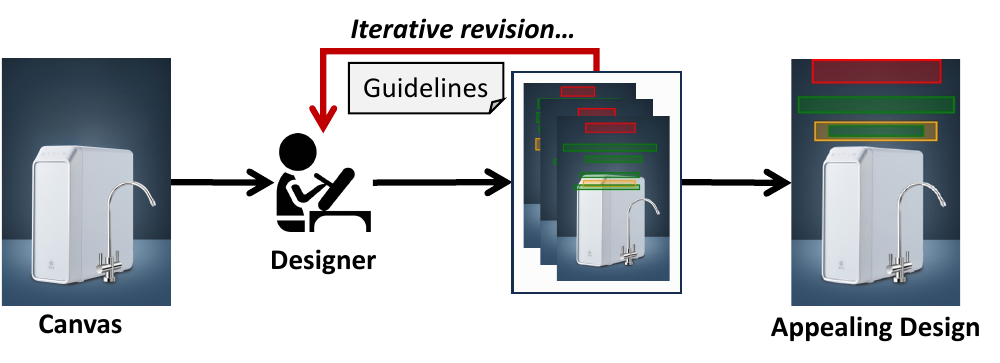}
    \vspace{-5mm}
    \caption{
      The iterative revision workflow of designers \cite{li2024revision, duan2024generating}.
    }
    \label{fig:motivation_b}
  \end{subfigure}
  \vspace{3mm}
  \begin{subfigure}{1\linewidth}
    \centering
    \includegraphics[width=\linewidth]{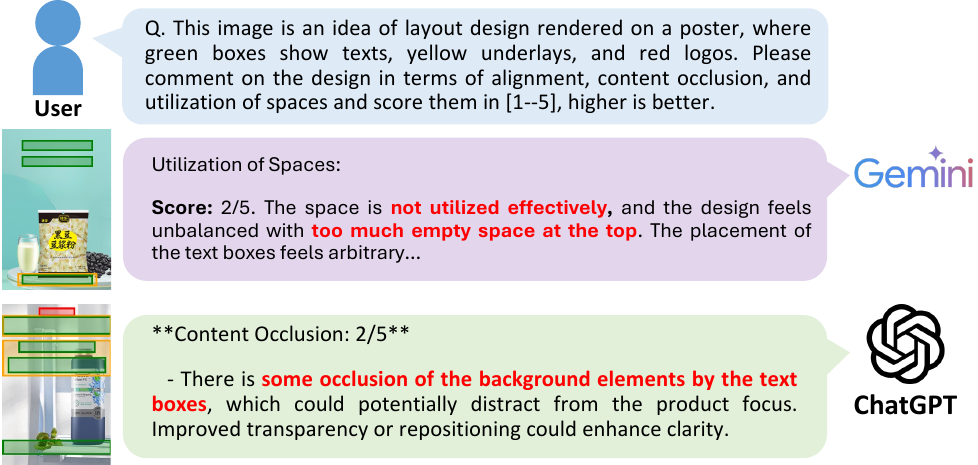}
    \caption{
      Inspiring conversations with Gemini/ChatGPT \cite{gemini, gpt4o}.
      }
    \label{fig:motivation_a}
  \end{subfigure}
  \vspace{-8mm}
  \caption{The designers' workflow follows an iterative revision process \cite{li2024revision} and heuristics based on guidelines \cite{duan2024generating}. The conversation highlights the ability of LVLMs to evaluate and identify design issues on \textit{rendered} problematic layouts, and provide valuable feedback. This capability is leveraged \textit{appropriately} to improve content-aware layout generation, like the workflow of designers.}
  \label{fig:motivation_combined}
  \vspace{-5mm}
\end{figure}

Content-aware layout generation is a kind of layout generation task~\cite{lok2001survey,shi2023intelligent} where layout elements are generated on the basis of a given visual content (\ie, canvas). It has emerged as a pivotal area of research with applications covering various tasks such as poster creation~\cite{guo2021vinci,hsu2023posterlayout} and magazine layouts~\cite{jahanian2013recommendation,yang2016automatic}. Usually, designers practice iterative processes \cite{li2024revision} and heuristics based on guidelines \cite{duan2024generating} to achieve visually appealing layouts (\cref{fig:motivation_combined}(a)), requiring significant expertise. However, emulating this process automatically is challenging due to the scarcity of high-quality, annotated datasets, which are difficult and costly to obtain.

Previous studies have attempted to automate content-aware layout generation by employing generative models, such as generative adversarial networks (GANs)~\cite{cgl-gan,hsu2023posterlayout},
Transformer-based models~\cite{horita2024retrieval}, and diffusion-based models~\cite{li2023relation,chai2023two}, to produce layouts based on visual content. While these methods have achieved
remarkable success, they often lack the iterative refinement process like human designers.
Some studies~\cite{lin2024spot, iwai2025layout} try to address this issue by introducing mechanism for iterative layout correction, yet these efforts still fall short due to the issue of limited data availability. Consequently, the layouts generated
by these models frequently require manual adjustments and fine-tuning by the designers to meet aesthetic standards. This highlights a significant gap in existing methodologies, where the ability to iteratively improve and refine layouts autonomously remains underdeveloped.

Recent advancements in large language models (LLMs) have demonstrated their remarkable
capacity to generate various forms of structured content, including layout designs, due to their vast knowledge base, which encompasses not only text but also design principles. By fine-tuning LLMs with extensive training data and computational resources, it has become possible to leverage their inherent understanding for layout generation~\cite{tang2024layoutnuwa,seol2024posterllama}.
Notably, proprietary LLMs have recently achieved impressive performance in this domain without additional training, using in-context learning (ICL)~\cite{gpt3} to generate layouts from only a few examples~\cite{lin2024layoutprompter}. These models have shown the ability to produce high-quality layouts, even when provided only with textual descriptions of layout elements. However, two key challenges remain in applying such methods to content-aware layout generation: 1) how to efficiently leverage visual content to enhance performance is unresolved, 
and 2) the aesthetic quality of generated layouts is not necessarily pleasing in subjective and objective evaluation.

To address these challenges, efficiently integrating multi-modal capabilities into layout generation is essential. However, transitioning the LLM's text-centric solution to environments that incorporate multiple modalities introduces new challenges in enhancing, evaluating, and validating multi-modal content, like sequences of interleaved images and text \cite{yang2024idea2img}. As shown in \cref{fig:motivation_combined}, inspired by the ability of large vision-language models (LVLMs) that can assess the quality of graphic design \cite{haraguchi2024can}, and the iterative design process \cite{li2024revision} following guidelines and heuristics \cite{nielsen1992finding, nielsen1990heuristic} commonly adopted by designers to create appealing layouts, we propose \textit{\textbf{\underline{V}}isual-\textbf{\underline{A}}ware \textbf{\underline{S}}elf-\textbf{\underline{C}}orrection L\textbf{\underline{A}}yout Gene\textbf{\underline{R}}ation} (\textbf{VASCAR}). VASCAR is a multi-modal \textit{training-free} framework for iteratively self-improving content-aware layout generation, designed to \textit{mimic designers' workflow} (\cref{fig:motivation_combined}(a)). 

Specifically, we propose using bounding box layouts rendered on canvas as visual prompting, since the findings show GPT-4V’s ability to understand a variety of markers \cite{yang2023dawn, yang2023set}. Given a canvas as a query and a few multi-modal ICL examples, we prompt the LVLM to generate multiple layout candidates in a \textit{parsable} HTML format. To mimic the design process, inspired by \cite{yang2024large}, we introduce a multi-modal self-correction method. The generated layouts are parsed and rendered on the query canvas as visual content. Additionally, we propose a layout scorer to assign unified scores as evaluation results and a suggester to provide recommendations for refining the generated layouts, mimicking established guidelines and heuristics. The optimized multi-modal prompt is then fed back to LVLM for another iteration, as self-refinement has been shown to improve performance in NLP tasks \cite{yang2024idea2img}. This process enables LVLMs to refine and enhance the layout design iteratively, much like a designer practicing to create aesthetical designs.

In experiments, we confirmed the effectiveness of VASCAR using the PKU and CGL poster-layout datasets~\cite{hsu2023posterlayout,cgl-gan}. On the relatively small PKU with 10k-order training samples, VASCAR outperformed all of the existing LLM-based and generative-model-based methods across most constraint-based metrics such as occlusion, readability, and distributional metrics, layout FID~\cite{kikuchi2021constrained}. In larger CGL, where layout-specific generative models enjoy their full modeling
abilities, VASCAR still outperformed them in most of the constraint-based metrics and marked competitive FID without training.

\textbf{Contributions.} We leverage LVLMs for content-aware layout generation in a \textit{training-free} manner for the first time. We propose a simple but effective method using bounding-box layouts rendered on canvas as visual prompting, serving as visual content for LVLMs. To mimic designers' workflow, we propose a layout scorer and a suggester to iteratively prompt LVLMs. Extensive experiments and user study show that VASCAR achieves the state-of-the-art (SOTA) content-aware layout generation results. The generalizability of VASCAR across LVLMs (\eg, GPT-4o and Gemini) demonstrates its versatility. 
Our findings on LVLMs' usability for content-aware layout generation serve as a step-stone for future research in advanced graphic design with LVLMs.

    \section{Related Work}
\label{sec:related}

\begin{figure*}[ht]
  \centering
  \includegraphics[width=\linewidth]{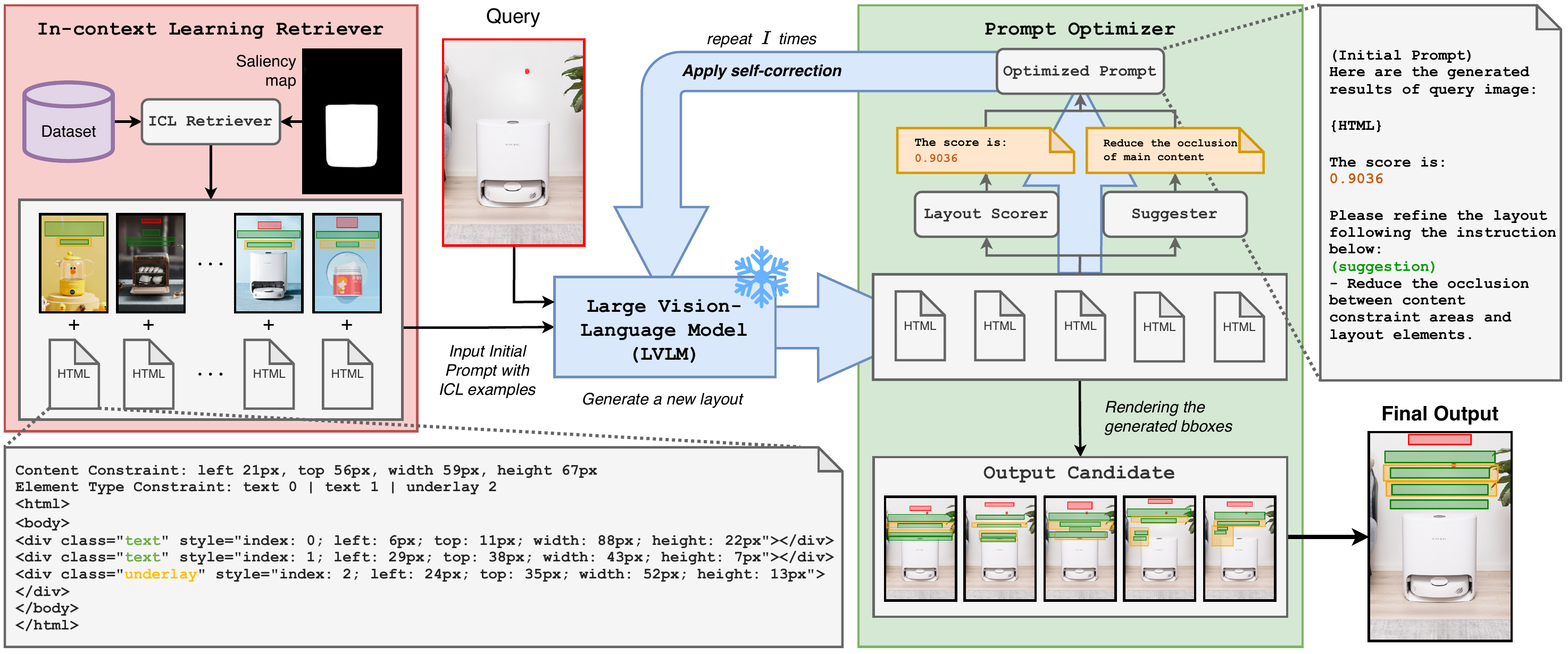}
  \vspace{-5mm}
  \caption{An overview of VASCAR. First,
  {\it ICL retriever} uses the saliency map of the query image to retrieve several ICL examples
  from the dataset. The retrieved samples are transformed to rendered images along
  with the text layout in HTML format. We input all the ICL examples and the
  query into a frozen LVLM to generate the layout in HTML format. The LVLM is queried iteratively with optimized prompts by {\it layout scorer} and {\it suggester}. After $I$ iterations, VASCAR return the candidate with the highest score.}
  \vspace{-4mm}
  \label{fig:overview}
\end{figure*}

\nbf{Layout-Specific Generative Models.} The exploration of layout generation~\cite{lok2001survey,shi2023intelligent}
began in a content-agnostic setting, initially employing optimization-based methods
guided by design principles~\cite{o2014learning,o2015designscape}. With the
advent of deep learning, methods based on generative adversarial networks (GANs)~\cite{li2019layoutgan,li2020attribute}
and variational auto-encoders (VAEs)~\cite{jyothi2019layoutvae,yamaguchi2021canvasvae}
were proposed, marking a significant shift in the field. Notably, optimizing
the latent variables of GANs to generate layouts, an approach known as CLG-LO~\cite{kikuchi2021constrained},
has demonstrated a certain efficacy in aesthetic constraints. Furthermore, LACE~\cite{chentowards}
demonstrated performance improvements by employing a post-processing method that
directly optimizes the constraints. Since then, Transformer~\cite{vaswani2017attention}-based
models~\cite{gupta2021layouttransformer,jiang2023layoutformer++}, and more
recently, diffusion-based~\cite{inoue2023layoutdm,chai2023layoutdm,iwai2025layout} and flow-based~\cite{guerreiro2025layoutflow} models
have been introduced to enhance the generative capabilities in content-agnostic scenarios.

Following the research in content-agnostic settings, content-aware layout
generation, which is more readily applicable in business contexts, has emerged, with
GAN-based~\cite{zheng2019content,cgl-gan,hsu2023posterlayout} and diffusion-based~\cite{li2023relation,chai2023two}
methods being proposed. These models are trained from scratch, but acquiring design
data is often more costly in terms of human and financial resources, leading to
a limited amount of data compared to other domains. As recent models become increasingly
data-hungry, the scarcity of training data remains a major challenge in this
field.

\nbf{Layout Generation by LLMs.} Recent studies pointed out that LLMs have
ability to generate layouts in the form of structure-description language such
as HTML~\cite{tang2024layoutnuwa,lin2024layoutprompter}. This might be surprising
because LLMs do not have explicit handling mechanisms for visual information,
but it is possible that LLMs have learned to solve layout as a programming task
\cite{austin2021program} from design-related sample codes of HTML contained in
the training corpora. This prior knowledge makes LLM-based layout generation promising,
especially in data-scarce scenarios. LayoutNUWA \cite{tang2024layoutnuwa} is one
of the pioneering work in this line by formulating layout generation as a code-completion
task using Code Llama~\cite{roziere2023code} language model. LayoutPromter
\cite{lin2024layoutprompter} and LayoutGPT \cite{feng2024layoutgpt} similarly
utilized GPT-3 \cite{gpt3} and showed that it can adapt to layout generation in
a training-free manner by using ICL. MuLCO~\cite{chen2024empowering} tackled multi-page
consistent layout generation with GPT. Although extending them with visual
modality is a natural direction, we are aware of only one existing study at present:
PoseterLlama \cite{seol2024posterllama} carefully fine-tuned a VLM \ie, MiniGPT-4~\cite{zhu2024minigpt}.
However, how to exploit proprietary LVLMs that we can not finetune for layout generation
is an open problem, and VASCAR is the first work to tackle this.

\nbf{Self-Correction in LLMs.} For exploiting full potentials of LLMs, various
prompting techniques are examined rather than simply breaking target tasks down
into questions and answers, which established a research area of prompt engineering~\cite{white2023prompt}.
Self-correction~\cite{welleck2023generating} is one of the promising directions among
them, where LLMs are tasked to correct mistakes in their own outputs with or without
external feedback. For example, feedback of search results from external corpora
is used in text-based question answering tasks~\cite{gao2023rarr}, execution
results are used as feedback in program synthesis~\cite{ni2023lever}, and object
detector is used for text-to-image generation~\cite{wu2024self}. Interestingly, even
without access to external information, LLMs can correct their own errors via
prompting. This is called intrinsic self-correction~\cite{madaan2023selfrefine},
although its efficacy is controversial \cite{huang2024large,kamoi2024can} in the
language processing area. In this context, VASCAR exploits two feedbacks for
self-correction: external feedback from automatic metrics and intrinsic multi-modal
feedback from rendered layout images, where the model's own textual output is
fed back in the form of visual input.
    \vspace{-2mm}
\section{Method}
\vspace{-2mm}

\Cref{fig:overview} shows an overview of VASCAR.
To achieve iterative self-corrective layout generation with an LVLM, our pipeline is based of the novel \textit{visual-aware self correction}, which consists our key contribution.
In each step, LVLM-based layout generation and refinement are performed.
VASCAR consists of \textit{layout generator} to generate the initial layouts, \textit{prompt optimizer} to evaluate the layouts based on automatic criteria and construct the next-step prompts, and \textit{ICL retriever} to identify multiple exemplars that serve as potential design constraints, each of whose details are described in the following sections.  

\subsection{Task Definition}
Let $\bm{x}_{\text{q}}$ denote the input canvas on which we generate a layout,
and $\mathcal{D}= \{(\bm{x}, \bm{y})\}$ the dataset of pairs of a canvas $\bm{x}$ and corresponding
annotated \textit{layout} $\bm{y}$, where $|\mathcal{D}| = N$. The canvas $\bm{x}$
may be an image of a product, an event, \etc, but without any visual layout elements.
A layout is a set of annotated visual elements in the materials, \ie,
$\bm{y} = \{\bm{e}\}$ where $|\bm{y}| = K$ and each visual element $\bm{e} = (c, l, t, w, h)$ is
represented by its \textit{element type} $c$ to specify the type of the visual element
(\eg, logo, text, and underlay) and the bounding box specified by the left-top-width-height coordinates. That is,
the arrangement of the visual elements in professionally designed material is
encoded in $\bm{y}$.\footnote{Note that $\bm{y}$ does not provide any information on the
content of each visual element $\bm{e} \in \bm{y}$. For example, if $e$ is a text block, the
text should be provided to fully re-generate the original material. 
We omit the content as VASCAR does not depend on it.} 
VASCAR generates a layout
$\bm{y}_{\text{q}}$ for $\bm{x}_{\text{q}}$ with referring to a subset of $\mathcal{D}$
identified by the ICL retriever. 

\subsection{Visual-Aware Self-Correction}
Visual-aware self-correction refers to our overall pipeline that first generates layouts on the query canvas, and refines the layouts iteratively with visual feedback.
First, we introduce a unified notation for prompt-based generation and refinement with an LLM or LVLM by
$
    Y_\text{q} = G(\bm{x}_\text{q}, Y; p),
$
where $G(\cdot)$ represents the generator, $\bm{x}_\text{q}$ denotes the input canvas, $Y$ denotes the current state of layouts under refinement if provided, and $p$ denotes a prompt text that instruct the model to generate or refine layouts.
The output $Y_\text{q}$ is a set of layouts each of whose elements is denoted by $\bm{y}_\text{q}$, since a generator may output multiple candidates of layouts per call.
The initial layout generation can be denoted as
\begin{equation}
    Y_\text{q}^0 = G(\bm{x}_\text{q}, \emptyset; p_0),
\end{equation} 
where $p_0$ is the default prompt that instructs the model to generate layouts from scratch.
To indicate the absence of the previous-step layouts, we explicitly use an empty set $\emptyset$.

To iteratively refine the layouts with self-feedback, the same LLM or LVLM used for generation can be employed by simply switching the prompt to a refinement mode. We construct refinement prompts by modifying the default prompt with additional instructions and the previous states of the generated layouts. This refinement step can be denoted by
\begin{equation}\label{label:lvlm}
    Y_\text{q}^{i} = G(\bm{x}_\text{q}, Y_\text{q}^{i - 1}; p(Y_\text{q}^{i - 1})),
\end{equation}
where $Y_\text{q}^{i}$ denotes layouts generated in the $i$-th iteration.
The prompt $p(Y_\text{q}^{i - 1})$ is a refinement prompt that is conditioned by the previous state $Y_\text{q}^{i - 1}$.

In constructing the function $G$ with an LVLM, we need to leverage its ability to access layout aesthetics by directly feeding a rough proxy of the resulting design with visual elements being given as their bounding boxes \cite{haraguchi2024can}, as shown in \cref{fig:motivation_combined}.  
We propose to feed these proxies, referred to as \textit{rendered images}, to the LVLM to provide richer cues for better layouts. 
A rendered image is generated by
placing bounding boxes with different colors for different element types, denoted by $\bm{z}(\bm{x}, \bm{y})$.
Most LVLMs accept a text prompt combined with multiple images, and we can concretize the function $G$ using an LVLM with a capacity for visual feedback by
\begin{equation} \label{eq:render}
    G(\bm{x}_\text{q}, Y; p) = \text{LVLM}(\{ \bm{z} (\bm{x}_\text{q}, \bm{y}_\text{q}) | \bm{y}_\text{q} \in Y \}, p),
\end{equation}
where $\text{LVLM}(V, t)$ denotes an LVLM viewed as a function that maps a pair of image set $V$ and prompt text $t$ on the output layout candidates.

Positive reinforcement in candidate-set feedback has been shown to be beneficial~ \cite{yang2024large} by keeping high-score examples for better results. Therefore, we select the top-5 layout from candidate pool using a scoring function $v(\bm{y}_{\text{q}})$ from $Y_\text{q}$ and discard the others to create the reference samples for the multi-modal optimized prompt. The initial prompt is also added into the multi-modal optimized prompt, as the LVLM needs to generate a parsable HTML layout. We perform self-correction for $I$ iterations by modifying the samples in the multi-modal optimized prompt, mimicking a designer’s iterative process, and select the layout with the highest score among $Y_\text{q}^I$ as the final output.

\subsection{Prompt Strategy for Layout Generator}
\label{sec:layout_generator}
Even for state-of-the-art LVLMs, generating properly formatted layouts remains challenging without exemplification. To address this, we construct our layout generator by a frozen LVLM enhanced with ICL \cite{lu2021fantastically, razeghi2022impact,
shin2022effect,zhang2024instruct}, incorporating a small subset of the training set into the prompts.
Given an input $\bm{x}_{\text{q}}$, we extract a small subset from the training set, denoted as $\mathcal{S}(\bm{x}_{\text{q}})$, where the subset size is ${M}$. These examples serve as ICL exemplars, allowing the layout generator to generate a layout $\bm{y}_{\text{q}}$ w.r.t. $\bm{x}_{\text{q}}$ and $\mathcal{S}(\bm{x}_{\text{q}})$.
The prompt $p_0$ consists of (i) instructions for generating a layout,
(ii) the input image $\bm{x}_{\text{q}}$, (iii) its saliency map $\bm{s}_{\text{q}}$, (iv)
rendered image $\bm{z}$ of each ICL example in $\mathcal{S}$, (v) saliency map $\bm{s}$ of
each ICL example, and (vi) the corresponding layout $\bm{y}$. Following \cite{lin2024layoutprompter}, the saliency map is represented as a serialized bounding box. Taking $\bm{s}_{\text{q}}$ as an example, it can be represented by a serialized bounding box $(l_{\text{q}}, t_{\text{q}}, w_{\text{q}}, h_{\text{q}})$, and is encoded into a textual format through the transformation $E_{\text{s}}$, \ie,
\begin{align} \label{label:constraint}
    E_{\text{s}}(s_{\text{q}}) = \; & \text{``\texttt{left $l_{\text{q}}$ px, top $t_{\text{q}}$ px,}}\nonumber     \\
                                    & \text{\texttt{width $w_{\text{q}}$ px, height $h_{\text{q}}$ px.}''}
\end{align}
With this default initial prompt $p_0$,
the LVLM generates multiple layouts $Y_{\text{q}}^0$, where different layouts can be generated by properly setting the temperature of the LVLM. 
Additionally, we reuse the rendered images defined in \cref{eq:render} for ICL samples to exploit LVLMs' multi-modal ability.
All rendered images from $\mathcal{S}(\bm{x}_{\text{q}})$ and the input image $\bm{x}_{\text{q}}$ are fed
into the LVLM alongside the rendered images for visual feedback.
In the self-correction steps, refinement prompts $p(Y_\text{q})$ follow the same structure as the initial prompt but differ in that they include the previous layout candidates, combined with textual \textit{refinement suggestion} as described in \cref{label:popt}, in addition to the ICL examples. 
The examples of real prompts can be found in Appendix.

\nbf{In-Context Learning Retriever}
Applying ICL to the visual domain requires selecting ICL samples that are relevant to the query image \cite{bar2022visual, zhang2024instruct, unsup}. We use the same retrieval rule as in previous work \cite{lin2024layoutprompter}, selecting samples with saliency maps~\cite{qin2019basnet,qin2022highly} similar to that of the query.
To efficiently reuse the saliency maps and images in \cref{label:constraint} and \cref{label:lvlm}, we extend the dataset $\mathcal{D}$ to include a saliency map for each pair, \ie, $\mathcal{D}' = \{(\bm{x}, \bm{y}, \bm{s}) \mid (\bm{x}, \bm{y}) \in \mathcal{D}\}$.


\vspace{-1mm}
\subsection{Multi-modal Prompt Optimizer}\label{label:popt}
 LVLMs' capability to access the suitability and aesthetics of generated layout is strong but still limited compared to fine-tuned models \cite{seol2024posterllama}. Generating multiple candidate layouts with ICL examples and choosing the best one can still be unsatisfactory \cite{lin2024layoutprompter}. Inspired by \cite{yang2024large}, and designers who iteratively create appropriate layouts for specific contexts and refine them based on guidelines \cite{duan2024generating, li2024revision}, we propose a multi-modal prompt optimizer. To achieve this, we introduce a \textit{layout scorer} to evaluate the generated layout and a \textit{suggester} to guide the LVLM in making effective adjustments.

The layout scorer uses multiple layout evaluation metrics (detailed in \cref{sec:evaluation_metrics}). Each metric is normalized to [0, 1] to fuse them,
where the larger value means better in terms of the criterion. Let $\mathcal{M}$
be the set of the criteria, and $f_{m}(\bm{x}_{\text{q}})$ gives the normalized score
of $\bm{y}_{\text{q}}$ for criterion $m \in \mathcal{M}$. The fused score is given by:
\begin{equation}
    v(\bm{y}_{\text{q}}) = \sum_{m \in \mathcal{M}}\lambda_{m}\cdot f_{m}(\bm{y}_{\text{q}}
    ),
\end{equation}
where $\lambda_{m}\in \lambda$ is an empirically determined weight.

The fused score $v(\bm{y}_{\text{q}})$ gives the layout generator ideas about how good
$\bm{y}_{\text{q}}$ is, but it does not tell which aspects are good and which are not.
Therefore, we propose a simple and effective suggester. The suggester uses individual
evaluation metric $f_{m}$. If it falls below a preset threshold, an additional instruction text is added, indicating the adjustment direction (\eg, ``\textit{Reduce the overlap}.'').
The threshold $\theta_{m}$ is set to the average of each evaluation metric across the ICL examples in $\mathcal{S}(s_{\text{q}})$, \ie,
\begin{align}
    \theta_{m}= \frac{1}{|\mathcal{S}(\bm{x}_{\text{q}})|}\sum_{\bm{y} \in \mathcal{S}(\bm{x}_\text{q})}f_{m}(\bm{y}).
\end{align}
The additional prompts from the layout scorer and suggester are then appended to the instruction in the refinement prompt $p(Y^i_\text{q})$.

    \begin{table*}
    [ht]
    \centering
    \resizebox{\textwidth}{!}{%
    \begin{tabular}{lccccccccccccc}
        \toprule                                                                       &                        & \multicolumn{6}{c}{PKU}     & \multicolumn{6}{c}{CGL}      \\
        \cmidrule(lr){3-8} \cmidrule(lr){9-14} \textbf{Method}                         & \textit{Training-free} & \multicolumn{2}{c}{Content} & \multicolumn{4}{c}{Graphic} & \multicolumn{2}{c}{Content} & \multicolumn{4}{c}{Graphic} \\
        \cmidrule(lr){3-4} \cmidrule(lr){5-8} \cmidrule(lr){9-10} \cmidrule(lr){11-14} &                        & Occ $\downarrow$            & Rea $\downarrow$            & Align $\downarrow$          & Und $\uparrow$             & Ove $\downarrow$ & FID$\downarrow$ & Occ $\downarrow$ & Rea $\downarrow$ & Align $\downarrow$ & Und $\uparrow$ & Ove $\downarrow$ & FID$\downarrow$ \\
        \midrule Real Data                                                             & -                      & 0.112                       & 0.0102                      & 0.0038                     & 0.99                       & 0.0009           & 1.58            & 0.125            & 0.0170           & 0.0024            & 0.98           & 0.0002           & 0.79            \\
        \midrule CGL-GAN \cite{cgl-gan}                                                & \xmark                 & 0.138                       & 0.0164                      & 0.0031                     & 0.41                       & 0.0740            & 34.51           & 0.157            & 0.0237           & 0.0032            & 0.29           & 0.1610           & 66.75           \\
        DS-GAN \cite{hsu2023posterlayout}                                              & \xmark                 & 0.142                       & 0.0169                      & 0.0035                     & 0.63                       & 0.0270            & 11.80           & 0.141            & 0.0229           & 0.0026            & 0.45           & 0.0570           & 41.57           \\
        ICVT \cite{icvt}                                                               & \xmark                 & 0.146                       & 0.0185                      & 0.0023                     & 0.49                       & 0.3180            & 39.13           & \textbf{0.124}   & 0.0205           & 0.0032            & 0.42           & 0.3100           & 65.34           \\
        LayoutDM \cite{inoue2023layoutdm}                                              & \xmark                 & 0.150                       & 0.0192                      & 0.0030                     & 0.41                       & 0.1900            & 27.09           & 0.127            & 0.0192           & 0.0024            & 0.82           & 0.0200           & 2.36            \\
        Autoreg \cite{horita2024retrieval}                                             & \xmark                 & 0.134                       & 0.0164                      & 0.0019                     & 0.43                       & 0.0190            & 13.59           & {\ul 0.125}      & 0.0190           & 0.0023            & 0.92           & 0.0110           & 2.89            \\
        RALF \cite{horita2024retrieval}                                                & \xmark                 & {\ul 0.119}                 & 0.0128                & 0.0027                     & 0.92                 & 0.0080      & 3.45            & {\ul 0.125}      & 0.0180           & 0.0024            & {\ul 0.98}  & 0.0040           & \textbf{1.32}   \\
        PosterLlama$^{\dagger}$ \cite{seol2024posterllama}                             & \xmark                 & --                          & --                          & --                          & --                         & --               & --              & 0.154            & {0.0135}     & {0.0008}       & 0.97     & 0.0030           & {\ul 2.21}      \\
        LayoutPrompter$^{\dagger}$ \cite{lin2024layoutprompter}                        & \checkmark             & 0.220                        & 0.0169                      & {\ul 0.0006}             & 0.91                       & {\ul 0.0003}  & {3.42}      & 0.251            & 0.0179           & \textbf{0.0004}    & 0.89           & \textbf{0.0002}  & 4.59            \\
        \rowcolor{LimeGreen!10} VASCAR (GPT-4o)                                        &     \checkmark         & 0.129                     & {\bf 0.0091}      &
        {\bf 0.0002}   & {\bf 0.99}                      & {\bf 0.0002}                     & {\bf 3.14}          & 0.141           & {\bf 0.0102}     & {\ul 0.0005}      & {\bf 0.99}          & {\bf 0.0002}           & 5.69          \\
        \rowcolor{RoyalBlue!10} VASCAR (Gemini)                                          & \checkmark             & \textbf{0.113}              & {\ul 0.0117}             & 0.0013                & {\ul 0.98}              & {\ul 0.0003}  & {\ul 3.34}   & {\ul 0.125}      & {\ul 0.0122}  & {0.0010}           & {\ul 0.98}  & {\ul 0.0007}     & 6.27            \\
        \bottomrule
    \end{tabular}
    }
    \vspace{-3.4mm}
    \caption{Unconstrained generation results on the PKU and CGL test split. Real
    Data is the test set with ground truth. $^{\dagger}$ shows our reproduced
    results based on the RALF split~\cite{horita2024retrieval}. Best and second-best
    results are \textbf{bold} and with {\ul underline}, respectively.}
    \vspace{-3mm}
    \label{table:unconstrained_results}
\end{table*}

\begin{table}[t]
    \centering
    \resizebox{1\columnwidth}{!}{%
    \begin{tabular}{lccccccccc}
        \toprule                                              & \multicolumn{4}{c}{PKU unannotated} & \multicolumn{4}{c}{CGL unannotated} \\
        \cmidrule(lr){2-5} \cmidrule(lr){6-9} \textbf{Method} & Occ $\downarrow$                    & Rea $\downarrow$                   & Und $\uparrow$ & Ove $\downarrow$ & Occ $\downarrow$ & Rea $\downarrow$ & Und $\uparrow$ & Ove $\downarrow$ \\
        \midrule CGL-GAN \cite{cgl-gan}                       & 0.191                               & 0.0312                             & 0.32           & 0.0690            & 0.481            & 0.0568           & 0.26           & 0.2690            \\
        DS-GAN \cite{hsu2023posterlayout}                     & 0.180                               & 0.0301                             & 0.52           & 0.0260            & 0.435            & 0.0563           & 0.29           & 0.0710            \\
        ICVT \cite{icvt}                                      & 0.189                               & 0.0317                             & 0.48           & 0.2920            & 0.446            & 0.0425           & 0.67           & 0.3010            \\
        LayoutDM \cite{inoue2023layoutdm}                     & 0.165                               & 0.0285                             & 0.38           & 0.2010            & 0.421            & 0.0506           & 0.49           & 0.0690            \\
        Autoreg \cite{horita2024retrieval}                    & 0.154                               & 0.0274                             & 0.35           & 0.0220            & 0.384            & 0.0427           & 0.76           & 0.0580            \\
        RALF \cite{horita2024retrieval}                       & {\ul 0.133}                         & {\ul 0.0231}                       & {\ul 0.87}     & {\ul 0.0180}      & {\ul 0.336}      & {\ul 0.0397}     & {\ul 0.93}     & {\ul 0.0270}      \\
        \rowcolor{RoyalBlue!10} VASCAR (Ours)                 & \textbf{0.132}                      & \textbf{0.0151}                    & \textbf{0.98}  & \textbf{0.0002}  & \textbf{0.289}   & \textbf{0.0300}  & \textbf{0.98}  & \textbf{0.0007}  \\
        \bottomrule
    \end{tabular}
    }
    \vspace{-3mm}
    \caption{Unconstrained generation results on the unannotated test split. Best and second-best results are \textbf{bold} and with {\ul underline}.}
    \vspace{-5mm}
    \label{table:unannotated_results}
\end{table}

\vspace{-2mm}
\section{Experiments}
\vspace{-1mm}

\subsection{Experimental Setup}
\vspace{-1mm}

\nbf{Datasets.} We conducted our experiments on two open-source datasets: PKU~\cite{hsu2023posterlayout} and CGL~\cite{cgl-gan}, both of which contain e-commerce posters featuring shopping product images. The PKU dataset includes three element types: \textit{logo}, \textit{text}, and \textit{underlay}. CGL provides an extra element type, \textit{embellishment}.
The PKU dataset consists of 9,974 annotated posters and 905 unannotated posters
(image-only). Meanwhile, CGL offers 60,548 annotated posters and 1,000
unannotated posters. To create image-layout pairs, following previous studies
\cite{cgl-gan, hsu2023posterlayout}, we used the state-of-the-art inpainting model \cite{suvorov2022resolution} to generate images devoid of any visual elements from the posters following \cite{horita2024retrieval}. Since neither dataset includes an official test split, we adhered to the dataset split and experimental framework in RALF \cite{horita2024retrieval}. RALF divided the data into 7,735/1,000/1,000 posters for training/validation/testing in PKU, and 48,544/6,002/6,002 for CGL. For evaluation, we retrieved ICL examples solely from the training set and evaluated on both annotated and unannotated posters.


\nbf{Comparisons.} We compare VASCAR with task-specific methods, such as CGL-GAN \cite{cgl-gan}, DS-GAN \cite{hsu2023posterlayout}, ICVT \cite{icvt}, LayoutDM
\cite{inoue2023layoutdm}, Autoreg and its extension RALF~\cite{horita2024retrieval};
with LLM-based layout generation, LayoutPrompter \cite{lin2024layoutprompter} and \textit{training-needed} multi-modal model PosterLlama \cite{seol2024posterllama};\footnote{PosterLlama needed re-training in our split, and we only could reproduce it in CGL, for which configuration files and
preprocessed data were provided in the official codebase.} and with Real Data (with ground-truth layout) for reference. We reproduced LayoutPrompter and PosterLlama's results due to differences in experimental setup (\eg, train/test splits).

\nbf{Evaluation Metrics.}
\label{sec:evaluation_metrics}
We adopt the following metrics from two perspectives as in previous studies
\cite{hsu2023posterlayout, horita2024retrieval}.
\begin{itemize} 
   \item 
    \textbf{Content metrics} consider the harmony between the generated
        layout $\bm{y}$ and the original image $\bm{x}$. \textit{Occlusion} (Occ $\downarrow$)
        calculates the overlap between the main visual (specified by $\bm{s}_{\text{q}}$)
        and all visual elements. \textit{Unreadability} (Rea $\downarrow$)
        evaluates the spatial gradients of $\bm{x}$ within all visual elements of the
        element type of \textit{text}.
    \item 
    \textbf{Graphic metrics} evaluate layout quality without factoring in
        the visual content on the canvas. \textit{Overlay} (Ove $\downarrow$)
        calculates the average IoU between all pairs of elements, excluding ones
        of element type \textit{underlay}. \textit{Non-alignment} (Align $\downarrow$)
        assesses the degree of spatial misalignment between elements
        \cite{li2020attribute}. \textit{Underlay Effectiveness} (Und $\uparrow$)
        evaluates the ratio of valid \textit{underlay} elements (\ie, the other elements
        are completely encompassed by the \textit{underlay} element) to total underlay
        elements. \textit{Fréchet Inception Distance} (FID $\downarrow$) measures
        the similarity between the distributions of real and generated layouts using
        a pre-trained model~\cite{kikuchi2021constrained}.
\end{itemize}

\nbf{Implementation Details.} \texttt{gemini-1.5-flash} and \texttt{gpt-4o} are used as the LVLMs.
We set the number $M$ of ICL examples to 10. The number of candidates (\ie, $|Y_{\text{q}}|$ is set to 5. The number $I$ of self-correction is set to 15 and 5 for Gemini and GPT-4o, respectively. Since GPT-3~\cite{gpt3}'s \texttt{text-davinci-003} used in LayoutPrompter~\cite{lin2024layoutprompter} is no longer available, we instead use \texttt{gpt-4o} for reproduction. We set the evaluation criteria for
the prompt optimizer as $\mathcal{M}= \{\text{Occ}, \text{Rea}, \text{Ove}, \text{Align}, \text{Und}\}$,\footnote{FID is
excluded because it is for a distribution but not a sample.} and the corresponding weights $\lambda = \{0.4, 0.4, 0.1, 0, 0.1\}$. Specifically, the weight for Align, $\lambda_{4}$, is set to 0 as the outputs were almost immune to it. The temperature for Gemini and GPT-4o is set to 1.4 and 0.7, respectively.

\vspace{-1mm}
\subsection{Input-unconstrained Generation}
\vspace{-1mm}

We followed the experimental setup of RALF~\cite{horita2024retrieval} to compare
ours with others over the input-unconstrained generation task in~\cref{table:unconstrained_results}. We observed that on PKU, VASCAR achieved the best performance across almost all metrics, significantly outperforming even fine-tuned baselines. In CGL, VASCAR also achieved competitive performance on Rea, Ove, and Und. In terms of FID, we found a large degradation, which we attribute to the fact that the model requires different numbers of self-corrections for easy and difficult samples. For simple samples, the LVLM can generate layouts that meet the metrics within the initial few iterations. However, as the iterations increase, the model begins to overly adjust to the suggestions from the suggester, leading to results that increasingly deviate from the true layout, thus causing the FID to worsen.
It is known that FID tends to perform better with slight noise added to the layout~\cite{otani2024ltsim}, and we believe that this deterioration in FID does not necessarily explain the quality of the generation.

\Cref{table:unannotated_results} shows the performance comparison on the unannotated test split of PKU and CGL. VASCAR outperformed all other models across all metrics. While all compared models experienced performance degradation on these unannotated splits compared to the annotated ones due to the domain gap between inpainted and clean canvases, VASCAR demonstrated remarkable robustness to this domain gap, particularly in terms of the Und and Ove metrics.

\begin{figure}[t]
    \centering
    \includegraphics[width=1\linewidth]{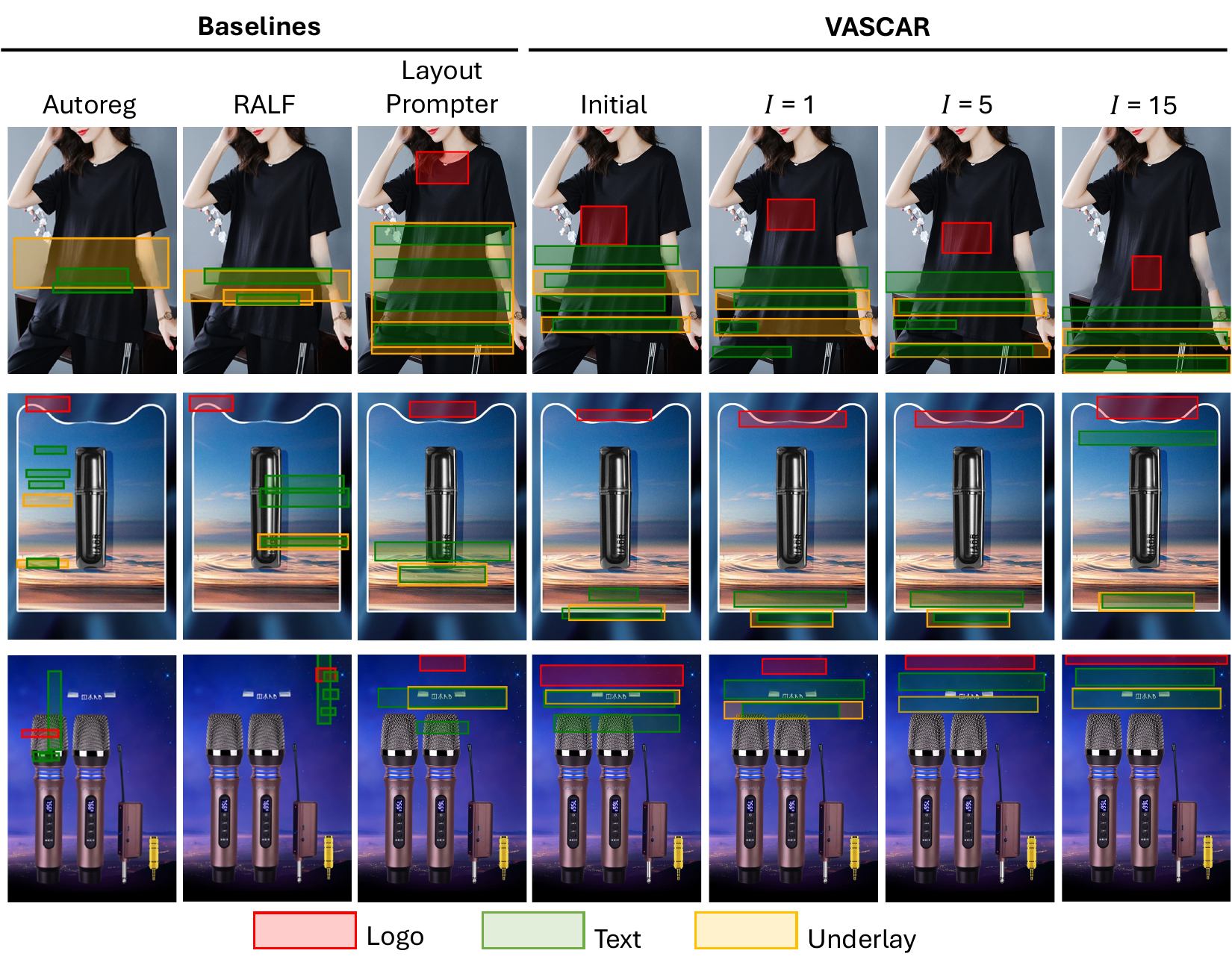}
    \vspace{-7mm}
    \caption{Visual comparison of baselines and VASCAR with different values of $I$. More examples can be found in Appendix.}
    \vspace{-7mm}
    \label{fig:visual_examples}
\end{figure}

\begin{figure*}[ht]
    \centering
    \begin{minipage}{0.6\textwidth}
        \centering
        \resizebox{\textwidth}{!}{%
        \begin{tabular}{lcccccccccc}
            \toprule                                                                                     & \multicolumn{5}{c}{PKU}     & \multicolumn{5}{c}{CGL}      \\
            \cmidrule(lr){2-6} \cmidrule(lr){7-11}                                                       & \multicolumn{2}{c}{Content} & \multicolumn{3}{c}{Graphic} & \multicolumn{2}{c}{Content} & \multicolumn{3}{c}{Graphic} \\
            \cmidrule(lr){2-3} \cmidrule(lr){4-6} \cmidrule(lr){7-8} \cmidrule(lr){9-11} \textbf{Method} & Occ $\downarrow$            & Rea $\downarrow$            & Und $\uparrow$              & Ove $\downarrow$           & FID$\downarrow$ & Occ $\downarrow$ & Rea $\downarrow$ & Und $\uparrow$ & Ove $\downarrow$ & FID$\downarrow$ \\
            \midrule Real Data                                                                           & 0.112                       & 0.0102                      & 0.99                        & 0.0009                     & 1.58            & 0.125            & 0.0170           & 0.98           & 0.0002           & 0.79            \\
            \midrule \textbf{C $\rightarrow$ S + P}                                                       \\
            CGL-GAN                                                                                      & 0.132                       & 0.0158                      & 0.48                        & 0.0380                     & 11.47           & 0.140            & 0.0213           & 0.65           & 0.0470            & 23.93           \\
            LayoutDM                                                                                     & 0.152                       & 0.0201                      & 0.46                        & 0.1720                     & 20.56           & 0.127            & 0.0192           & 0.79           & 0.0260            & 3.39            \\
            Autoreg                                                                                      & 0.135                       & 0.0167                      & 0.43                        & 0.0280                     & 10.48           & {\ul 0.124}      & 0.0188           & 0.89           & 0.0150            & {\ul 1.36}      \\
            RALF                                                                                         & 0.124                 & 0.0138                      & 0.90                        & 0.0100                     & \textbf{2.21}            & 0.126            & 0.0180           & 0.97     & 0.0060            & \textbf{0.50}   \\
            \rowcolor{LimeGreen!10} VASCAR (GPT-4o)                                                       & {\ul 0.117}                       & {\bf 0.0094}                & {\bf 1.00}                  & {\bf 0.0002}               & {3.01}   & 0.139            & {\bf 0.0099}     & {\bf 0.99}  & {\bf 0.0002}     & 4.86            \\
            \rowcolor{RoyalBlue!10} VASCAR (Gemini)                                                      & \textbf{0.107}              &        {\ul 0.0100}             & {\ul 0.99}               &           {\ul 0.0003}            & {\ul 2.82}      & \textbf{0.123}   &  {\ul 0.0111}  & {\ul 0.98}  & {\ul 0.0005}  & 5.51            \\
            \midrule \textbf{C + S $\rightarrow$ P}                                                       \\
            CGL-GAN                                                                                      & 0.129                       & 0.0155                      & 0.48                        & 0.0430                     & 9.11            & 0.129            & 0.0202           & 0.75           & 0.0270            & 6.96            \\
            LayoutDM                                                                                     & 0.143                       & 0.0185                      & 0.45                        & 0.1220                     & 24.90           & {\ul 0.127}      & 0.0190           & 0.82           & 0.0210            & 2.18            \\
            Autoreg                                                                                      & 0.137                       & 0.0169                      & 0.46                        & 0.0280                     & 5.46            & {\ul 0.127}      & 0.0191           & {\ul 0.88}     & 0.0130            & {\ul 0.47}      \\
            RALF                                                                                         & 0.125                 & 0.0138                      & 0.87                  & 0.0100                     & \textbf{0.62}   & 0.128            & 0.0185           & \textbf{0.96}  & 0.0060            & \textbf{0.21}   \\
            \rowcolor{LimeGreen!10} VASCAR (GPT-4o)                                                       & {\ul 0.123}                 & {\ul 0.0117}                & {\bf 0.90}                        & {\bf 0.0009}               & {\ul 1.11}      & 0.140            & {\ul 0.0132}     & {\ul 0.88}           & {\bf 0.0003}     & 1.67            \\
            \rowcolor{RoyalBlue!10} VASCAR (Gemini)                                                      & \textbf{0.104}              & \textbf{0.0107}             & {\ul 0.88}               &        {\ul 0.0018}            & 2.21            & \textbf{0.122}   & \textbf{0.0123}  & 0.85           & {\ul 0.0028}  & 2.39            \\
            \midrule \textbf{Completion}                                                                  \\
            CGL-GAN                                                                                      & 0.150                       & 0.0174                      & 0.43                        & 0.0610                     & 25.67           & 0.174            & 0.0231           & 0.21           & 0.1820           & 78.44           \\
            LayoutDM                                                                                     & 0.135                       & 0.0175                      & 0.35                        & 0.1340                     & 21.70           & 0.127            & 0.0192           & 0.76           & 0.0200           & 3.19            \\
            Autoreg                                                                                      & 0.125                       & 0.0161                      & 0.42                        & 0.0230                     & 5.96            & \textbf{0.124}   & 0.0185           & 0.91           & 0.0110           & {\ul 2.33}      \\
            RALF                                                                                         & {\ul 0.120}                 & 0.0140                      & 0.88                  & 0.0120                     & \textbf{1.58}   & {\ul 0.126}      & 0.0185           & 0.96           & 0.0050           & \textbf{1.04}   \\
            \rowcolor{LimeGreen!10} VASCAR (GPT-4o)                                                       & {\ul 0.120}                       & {\bf 0.0063}                & \textbf{1.00}               & {\bf 0.0004}               & {6.19}      & 0.137            & {\bf 0.0068}     & {\bf 0.99}     & {\bf 0.0005}     & 5.57            \\
            \rowcolor{RoyalBlue!10} VASCAR (Gemini)                                                      & \textbf{0.119}              &        {\ul 0.0097}             & {\ul 0.99}               & {\ul 0.0006}            & {\ul 4.74}            & 0.135            & {\ul 0.0098}  & {\ul 0.98}  & {\ul 0.0009}  & 5.18            \\
            \midrule \textbf{Refinement}                                                                  \\
            CGL-GAN                                                                                      & 0.122                       & 0.0141                      & 0.39                        & 0.0900                      & 6.40            & {\ul 0.124}      & 0.0182           & 0.86           & 0.0240           & 1.20            \\
            LayoutDM                                                                                     & 0.115                       & 0.0174                      & 0.40                        & 0.0630                      & 2.86            & 0.127            & 0.0196           & 0.75           & 0.0150           & 1.98            \\
            Autoreg                                                                                      & 0.131                       & 0.0171                      & 0.40                        & 0.0220                      & 5.89            & 0.126            & 0.0183           & 0.89           & 0.0040           & {\ul 0.15}      \\
            RALF                                                                                         & 0.113                 & 0.0109                      & 0.95                  & 0.0040                      & \textbf{0.13}   & 0.126            & 0.0176           & \textbf{0.98}  & 0.0020     & \textbf{0.14}   \\
            \rowcolor{LimeGreen!10} VASCAR (GPT-4o)                                                       & {\ul 0.108}                       & {\bf 0.0072}                & {\bf 0.99}                  & {\bf 0.0005}                & {1.18}      & 0.125      & {\bf 0.0091}     & 0.95           & {\bf 0.0008}           & 0.59            \\
            \rowcolor{RoyalBlue!10} VASCAR (Gemini)                                                      & \textbf{0.104}              &        {\ul 0.0095}             & {\ul 0.97}               &           {\ul 0.0010}             & {\ul 0.32}            & \textbf{0.114}   & {\ul 0.0096}  & {\ul 0.96}     & {\ul 0.0013}  & 0.37            \\
            \midrule \textbf{Relationship}                                                                \\
            Autoreg                                                                                      & 0.140                       & 0.0177                      & 0.44                        & 0.0280                     & 10.61           & {\ul 0.127}      & 0.0189           & 0.88           & 0.0150           & {\ul 1.28}      \\
            RALF                                                                                         & {\ul 0.122}                 & 0.0141                      & 0.85                        & 0.0090                     & 2.23            & \textbf{0.126}   & 0.0184           & \textbf{0.95}  & 0.0060           & \textbf{0.55}   \\
            \rowcolor{LimeGreen!10} VASCAR (GPT-4o)                                                       & 0.151                       & {\ul 0.0139}                & {\ul 0.92}                  & {\ul 0.0011}               & \textbf{1.94}   & 0.153            & {\ul 0.0139}     & {\ul 0.91}     & {\bf 0.0004}     & 2.61            \\
            \rowcolor{RoyalBlue!10} VASCAR (Gemini)                                                      & \textbf{0.119}              & \textbf{0.0117}             & \textbf{0.96}               & \textbf{0.0008}            & {\ul 2.00}      & 0.132            & \textbf{0.0123}  & \textbf{0.95}  & {\ul 0.0011}  & 3.72            \\
            \bottomrule
        \end{tabular}
        } \vspace{-3mm} \captionof{table}{Quantitative result of five constrained generation tasks on the PKU and CGL test splits. Best and second-best results are in \textbf{bold} and {\ul underline}, respectively.}
        \label{table:constrained_results}
    \end{minipage}
    \hfill
    \begin{minipage}{0.38\textwidth}
        \begin{minipage}{\textwidth}
        \centering
        \includegraphics[width=0.98\textwidth]{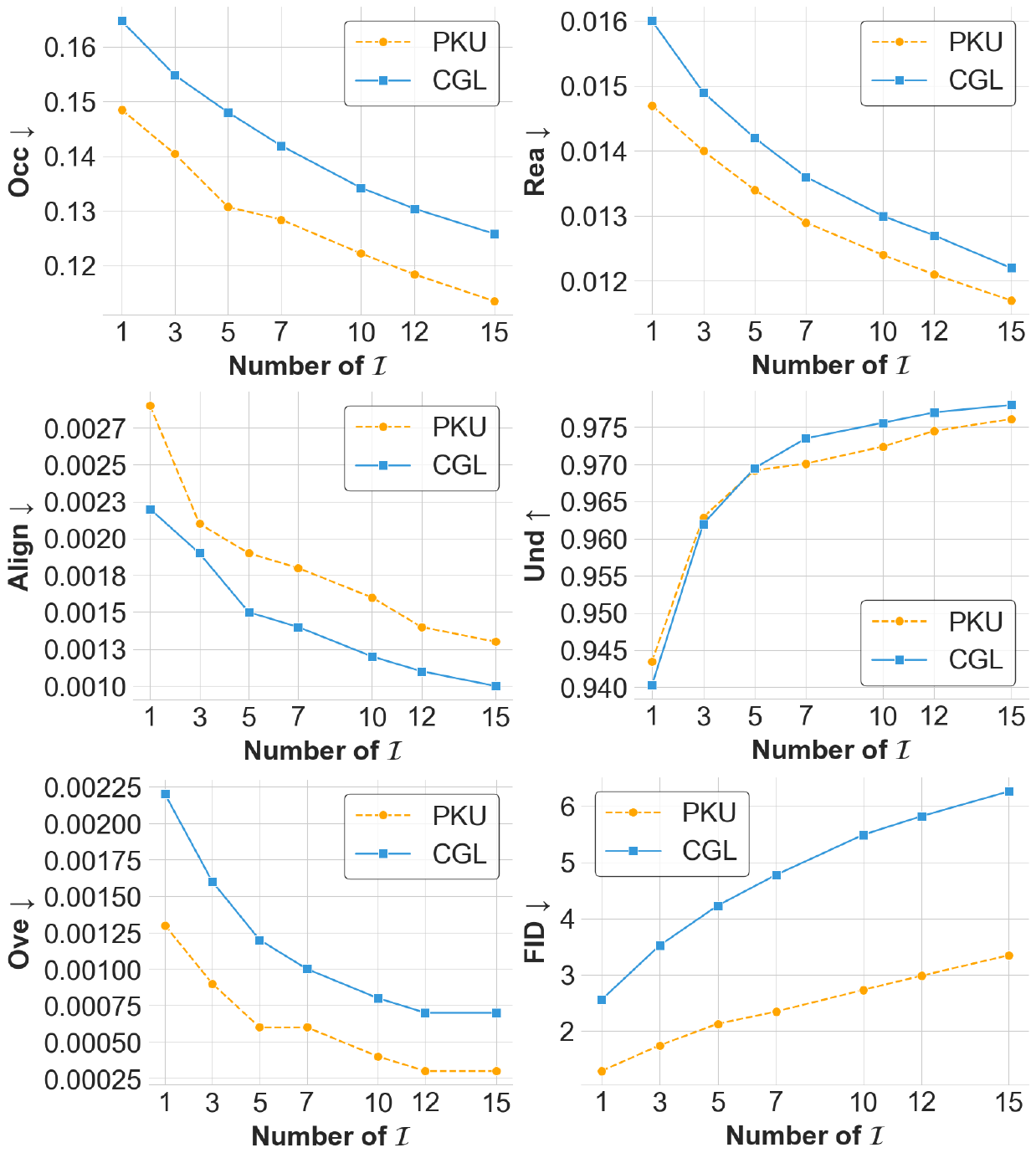}
        \vspace{-4mm}
        \captionof{figure}{The trends of each metric based on the number of self-correction $I$.}
        \label{fig:sc_trend}
        \end{minipage}
        \begin{minipage}{\textwidth}
        \centering
        \includegraphics[width=0.9\textwidth]{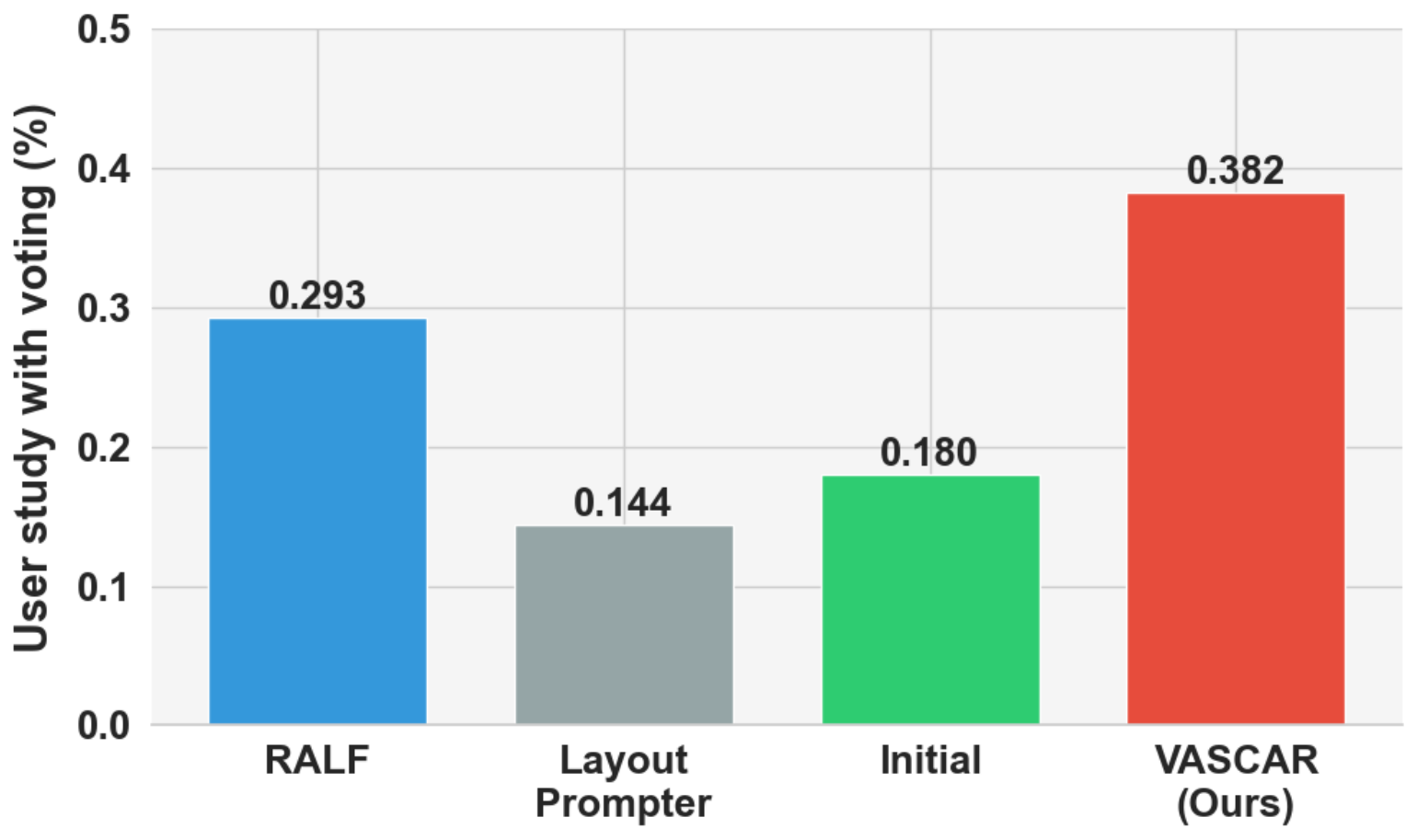}
        \vspace{-4mm}
        \captionof{figure}{User study results on PKU test split.}
        \label{fig:user_study}
        \end{minipage}
    \end{minipage}
    \vspace{-5mm}
\end{figure*}

\nbf{Qualitative Comparison.} 
\Cref{fig:visual_examples} shows the visual comparison of several baselines including Autoreg, RALF, and LayoutPrompter vs. our VASCAR across different self-correction $I$ including Initial \ie $I=0$. We confirm that VASCAR can generate well-arranged layouts avoiding overlapping and occluding the main content. 
VASCAR's results improve progressively with iterations.

\nbf{Trends of each metric.}
As shown in \cref{fig:sc_trend}, all metrics improve as the number of self-correction iterations increases, except for FID.

\nbf{Human evaluation.}
To further demonstrate the effectiveness of VASCAR, we conducted a user study on the PKU test split. comparing it against strong baselines: RALF \cite{horita2024retrieval}, LayoutPrompter \cite{lin2024layoutprompter}, and Initial. We invited 15 participants to evaluate 30 groups, selecting one preferred choice from each group. The user study results, based on voting percentages, are shown in \cref{fig:user_study}. VASCAR outperforms other baselines significantly according to human evaluation.

\vspace{-2mm}
\subsection{Input-constrained Generation}
\vspace{-2mm}


We evaluated VASCAR on the five input-constrained generation tasks followed by
Horita \etal \cite{horita2024retrieval}:

\nbf{Category $\rightarrow$ Size + Position} (C $\rightarrow$ S + P) task
specifies the category and the number of visual elements to place.

\nbf{Category + Size $\rightarrow$ Position} (C + S$\rightarrow$P) specifies the
size of elements to generate in addition to the C $\rightarrow$ S + P task.

\nbf{Completion} refers to a task where the model is given all information about
partial element (including category, size, and position) and asked to complete
the layout.

\nbf{Refinement} represents the task of improving a perturbed layout, which is
generated by adding Gaussian noise of mean 0 and variance 1 to the ground-truth
layout~\cite{rahman2021ruite}.

\nbf{Relationship} specifies the positional and size relationships between elements.

\begin{table}[t]
    \centering
    \resizebox{\columnwidth}{!}{%
    \begin{tabular}{lcccccc}
        \toprule                                               & \multicolumn{2}{c}{Content} & \multicolumn{4}{c}{Graphic} \\
        \cmidrule(lr){2-3} \cmidrule(lr){4-7} \textbf{Setting} & Occ $\downarrow$            & Rea $\downarrow$           & Align $\downarrow$ & Und $\uparrow$ & Ove $\downarrow$ & FID$\downarrow$ \\
        \midrule \rowcolor{RoyalBlue!10} Rendered Image (Ours) & \textbf{0.1304}             & \textbf{0.0134}            & \textbf{0.0017}    & 0.97           & 0.0012           & 2.13            \\
        \midrule
        Text-only                                              & 0.1529                      & 0.0153                     & 0.0024             & 0.97           & 0.0009           & 1.91            \\
        \midrule
        Saliency Map                                           & 0.1356                      & 0.0140                     & 0.0023             & 0.97           & 0.0009           & \textbf{1.69}   \\
        Inpainting Image                                       & 0.1312                      & 0.0137                     & 0.0022             & \textbf{0.98}  & \textbf{0.0004}  & 2.37            \\
        Original Poster                                        & 0.1315                      & \textbf{0.0134}            & 0.0023             & \textbf{0.98}  & 0.0006           & 2.34            \\
        \bottomrule
    \end{tabular}%
    }
    \vspace{-3mm}
    \caption{Multi-modal analysis for VASCAR on the PKU test split.}
    \vspace{-7mm}
    \label{table:multi_modal}
\end{table}

We adopted the text prompt preprocessing inspired by \cite{lin2024layoutprompter}, allowing VASCAR to handle these tasks suitable for content-aware layout generation. The details of the prompt can be found in Appendix.

As shown in \cref{table:constrained_results}, we found that VASCAR performs well
across several input-constrained generation tasks. On PKU, VASCAR even outperforms the ground-truth layout (Real Data in the table) on certain metrics, proving its effectiveness. We observed that VASCAR's FID scores are lower than those of RALF on some tasks, which we consider is related to the weight $\lambda_{m}$ in the layout scorer; excessive emphasis on the Occ and Rea metrics increases the distance between the generated layout and the ground-truth layout. This is further evidenced by its significantly better performance on content-related metrics compared to the ground-truth layouts.

\vspace{-2mm}
\subsection{Ablation Study}
\vspace{-2mm}
This section studies the impact of our design choices on PKU test split. Due to API limitations, $I$ is set to 5.

\nbf{The impact of multi-modal input.} We argue that the visual content determines the performance of layout generation model. Therefore, we compared using only text input and different types of image input (the same image type was also used in self-correction). \Cref{table:multi_modal} shows that using only text input shows a significant gap in content metrics compared to multi-modal input. Regarding different image types, rendered images achieved the best performance in content metrics and comparable performance in graphic metrics, demonstrating that rendered images enhance LVLMs' ability to generate layouts effectively.


\nbf{Impact of the number of ICL examples.}
The number of ICL examples plays a crucial role in the performance of LVLMs. When $M = 0$ (zero-shot), the models often fail to produce reasonable results \cite{lin2024layoutprompter}. To explore this effect, we evaluate VASCAR using different numbers of ICL examples. The results are summarized in \cref{table:combined_analysis}. Specifically, we experiment with $M = 1$, $3$, $5$, and $10$. Our findings indicate that as $M$ increases, performance improves across almost all metrics, with the exception of FID and Ove.

\begin{table}[t]
    \centering
    \resizebox{\columnwidth}{!}{%
    \begin{tabular}{lcccccc}
        \toprule
        & \multicolumn{2}{c}{Content} & \multicolumn{4}{c}{Graphic} \\
        \cmidrule(lr){2-3} \cmidrule(lr){4-7}
        \textbf{Setting} & Occ $\downarrow$ & Rea $\downarrow$ & Align $\downarrow$ & Und $\uparrow$ & Ove $\downarrow$ & FID$\downarrow$ \\
        \midrule
        \multicolumn{7}{l}{\textbf{Number of ICL Examples ($M$)}} \\
        1 & 0.2014 & 0.0184 & 0.0026 & 0.84 & 0.0017 & \textbf{1.45} \\
        3 & 0.1617 & 0.0160 & 0.0020 & 0.90 & 0.0016 & 2.27 \\
        5 & 0.1466 & 0.0150 & 0.0018 & 0.95 & \textbf{0.0006} & 2.25 \\
        \rowcolor{RoyalBlue!10} 10 (Ours) & \textbf{0.1304} & \textbf{0.0134} & \textbf{0.0017} & \textbf{0.97} & 0.0012 & 2.13 \\
        \midrule
        \multicolumn{7}{l}{\textbf{Number of Output Candidates ($|Y_{\text{q}}|$)}} \\
        1 & 0.1659 & 0.0180 & 0.0023 & 0.89 & 0.0011 & 2.29 \\
        3 & 0.1393 & 0.0150 & \textbf{0.0017} & 0.96 & 0.0010 & 2.24 \\
        \rowcolor{RoyalBlue!10} 5 (Ours) & 0.1304 & 0.0134 & \textbf{0.0017} & \textbf{0.97} & 0.0012 & 2.13 \\
        10 & \textbf{0.1225} & \textbf{0.0127} & 0.0025 & \textbf{0.97} & \textbf{0.0008} & \textbf{2.04} \\
        \midrule
        \multicolumn{7}{l}{\textbf{Ablation Study on $\lambda$}} \\
        \rowcolor{RoyalBlue!10} VASCAR (Ours) & 0.1304 & 0.0134 & \textbf{0.0017} & \textbf{0.97} & 0.0012 & 2.13 \\
        Initial & 0.2080 & 0.0218 & 0.0040 & \textbf{0.97} & 0.0013 & 1.05 \\
        w/o Occ & {\color{red}0.1709} & \textbf{{\color{RoyalBlue}0.0114}} & {\color{red}0.0022} & \textbf{0.97} & \textbf{{\color{RoyalBlue}0.0002}} & \textbf{{\color{RoyalBlue}1.95}} \\
        w/o Rea & {\color{RoyalBlue}0.1284} & {\color{red}0.0162} & {\color{red}0.0024} & \textbf{0.97} & {\color{RoyalBlue}0.0008} & {\color{RoyalBlue}2.03} \\
        w/o Ove & {\color{RoyalBlue}0.1301} & {\color{red}0.0137} & {\color{red}0.0018} & \textbf{0.97} & {\color{red}0.0030} & {\color{red}2.23} \\
        w/o Und & \textbf{{\color{RoyalBlue}0.1237}} & {\color{RoyalBlue}0.0130} & {\color{red}0.0023} & {\color{red}0.68} & {\color{RoyalBlue}0.0007} & {\color{red}2.14} \\
        \bottomrule
    \end{tabular}%
    }
    \vspace{-3mm}
    \caption{Comparison of results across various experimental settings on the PKU test split. ``Initial'' means without self-correction. ``w/o'' means without. The {\color{RoyalBlue}blue} and {\color{red}red} values indicate improved and degraded metrics compared to VASCAR.}
    \vspace{-5mm}
    \label{table:combined_analysis}
\end{table}

\nbf{Impact of the number of output candidates.}
We examine the effect of the number of output candidates ($|Y_{\text{q}}|$) in \cref{table:combined_analysis}. Our findings indicate that as $|Y_{\text{q}}|$ increases, VASCAR's performance improves accordingly. Considering the trade-off between performance and efficiency, we set $|Y_{\text{q}}| = 5$ as the default setting for VASCAR.

\nbf{Impact of $\lambda$ on the layout scorer.}
We evaluate the impact of the hyperparameter $\lambda$ on the layout scorer by the following configurations: (1) \textit{Without Occ}, (2) \textit{Without Rea}, (3) \textit{Without Ove}, and (4) \textit{Without Und}. The results are summarized in \cref{table:combined_analysis}. Our findings reveal that omitting a specific $\lambda$ for the layout scorer leads to a significant degradation in the corresponding metric. While other metrics may show slight improvements in some cases, they remain optimized compared to the ``Initial.'' Therefore, directly optimizing specific metrics is a reasonable approach. We also include VASCAR’s strength compared to the gradient-based LACE \cite{chentowards}, which is optimized for specific metrics in Appendix.

\begin{figure}[t]
    \centering
    \includegraphics[width=0.95\linewidth]{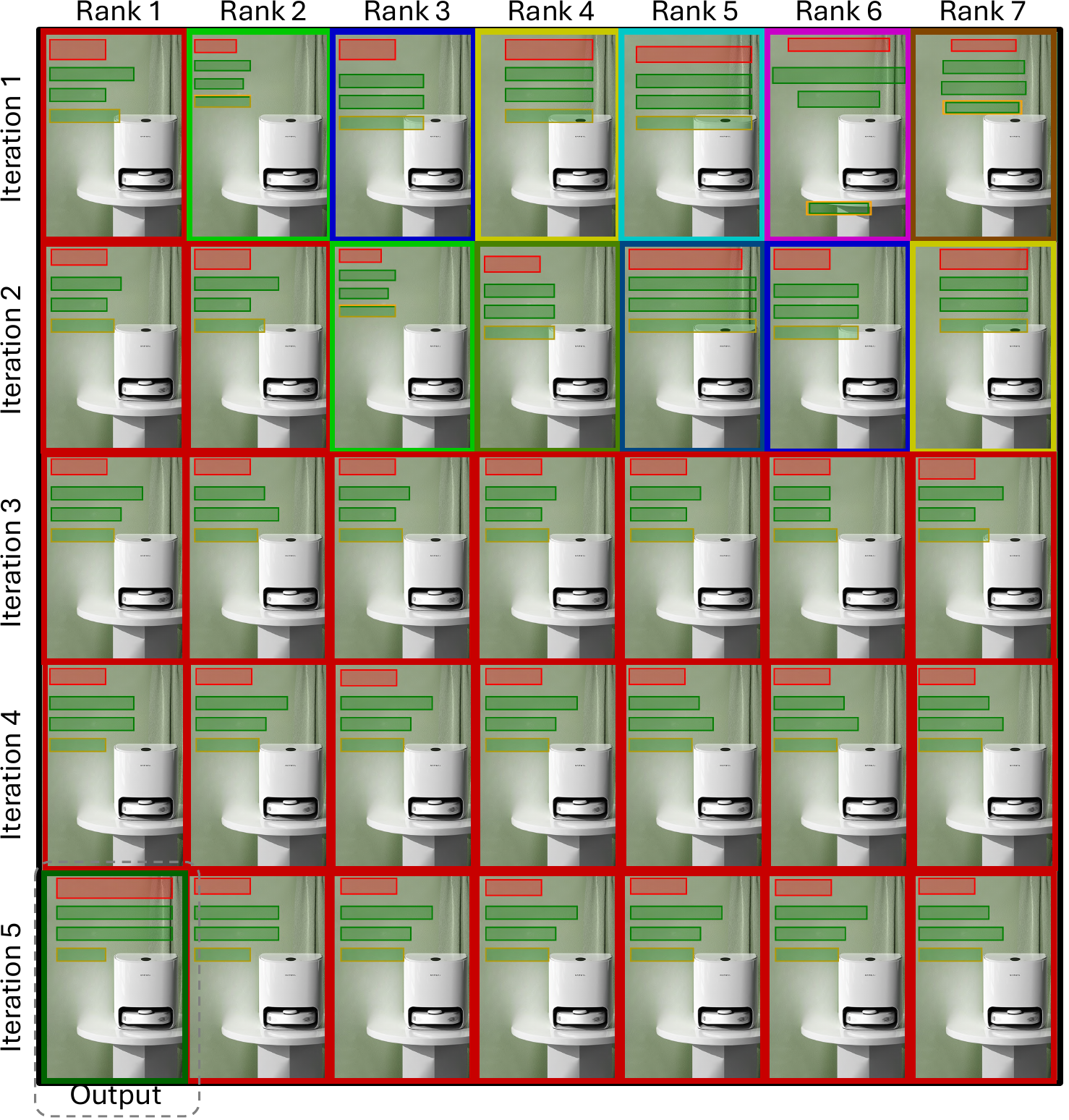}
    \vspace{-3mm}
    \caption{Visualization of the candidate pools under self-correction process
    with VASCAR. Best viewed in color.}
    \vspace{-6mm}
    \label{fig:correction}
\end{figure}


\nbf{In-depth analysis.}
\Cref{fig:sc_trend} shows that as the number of self-correction increases, all metrics improve except for FID, demonstrating VASCAR's effectiveness.
\Cref{fig:correction} shows an example with multiple candidate layout (\ie,
Rank 1 to Rank 7). The colored frame of each poster indicates identical or
similar layouts grouped by clustering \cite{ester1996density} based on IoU among
layouts; the same-colored frames mean that they are in the same cluster. In
earlier steps, the layouts show diversity and converge into similar or near-duplicate layouts after some iterations.

    \vspace{-2mm}
\section{Conclusion}
\vspace{-2mm}


This paper introduces VASCAR, a novel \textit{training-free} framework for content-aware layout generation that utilizes LVLMs through a visual-aware self-correction mechanism like a designer. The method iteratively refines layouts by incorporating feedback from both rendered visual contents and automatic metrics, improving layout quality without additional training. Extensive experiments and user study demonstrate that VASCAR excels in generating high-quality layouts and adapts well to different datasets and different LVLMs, offering a versatile solution for content-aware layout generation tasks.

\nbf{Limitations.} In exchange for improved layout quality, VASCAR needs more API
calls and token consumption than other single-shot API-based methods, resulting
in consumption of more money and time for generation. However, the consumption
is affordable at the moment; VASCAR costs less than 100 USD and 1.5 hours for 1,000-sample generation for the PKU experiments.

    { \small \bibliographystyle{ieeenat_fullname} \bibliography{main} }

    \clearpage
\setcounter{page}{1}
\maketitlesupplementary

\begin{table}[t]
    \centering
    \resizebox{\columnwidth}{!}{%
    \begin{tabular}{lcccccc}
        \toprule                                               & \multicolumn{2}{c}{Content} & \multicolumn{4}{c}{Graphic} \\
        \cmidrule(lr){2-3} \cmidrule(lr){4-7} \textbf{Setting} & Occ $\downarrow$            & Rea $\downarrow$           & Align $\downarrow$        & Und $\uparrow$  & Ove $\downarrow$          & FID$\downarrow$   \\
        \midrule Initial                                       & 0.210                       & 0.0218                     & 0.0040                    & 0.97            & 0.0013                    & 1.05             \\
        Initial + LACE                                         & \color{RoyalBlue}0.209      & \color{red}0.0233          & \color{RoyalBlue}0.0034   & \color{red}0.44 & \color{RoyalBlue}0.0002   & \color{red}1.06  \\
        \rowcolor{RoyalBlue!10} VASCAR (Ours)                  & {\color{RoyalBlue}0.130}    & {\color{RoyalBlue}0.0134}  & {\color{RoyalBlue}0.0017} & {0.97}          & {\color{RoyalBlue}0.0012} & {\color{red}2.13} \\
        \bottomrule
    \end{tabular}%
    }
    \caption{Comparison of post-hoc constraint optimization with LACE~\cite{chentowards}
    and our VASCAR on the PKU test split. {\color{RoyalBlue}Blue} and {\color{red}red}
    numbers show improved and degraded metrics compared to Initial.}
    \label{table:lace}
\end{table}

\begin{table}[t]
    \centering
    \resizebox{\columnwidth}{!}{%
    \begin{tabular}{lcccccc}
        \toprule                                               & \multicolumn{2}{c}{Content} & \multicolumn{4}{c}{Graphic} \\
        \cmidrule(lr){2-3} \cmidrule(lr){4-7} \textbf{Setting} & Occ $\downarrow$            & Rea $\downarrow$           & Align $\downarrow$ & Und $\uparrow$ & Ove $\downarrow$ & FID$\downarrow$ \\
        \midrule \rowcolor{RoyalBlue!10} IoU (Ours)            & \textbf{0.1304}             & \textbf{0.0134}            & \textbf{0.0017}    & \textbf{0.97}  & 0.0012           & \textbf{2.13}   \\
        \midrule Random                                        & 0.1740                      & 0.0178                     & 0.0018             & \textbf{0.97}  & \textbf{0.0002}  & 10.72            \\
        DreamSim \cite{fu2024dreamsim}                         & 0.1603                      & 0.0149                     & 0.0018             & 0.96           & 0.0010           & 3.77            \\
        \bottomrule
    \end{tabular}%
    }
    \caption{Comparison of results under different settings for ICL example
    selection on the PKU test split.}
    \label{table:icl_selection}
\end{table}

\begin{table}[t]
    \centering
    \resizebox{\columnwidth}{!}{%
    \begin{tabular}{lcccccc}
        \toprule                                               & \multicolumn{2}{c}{Content} & \multicolumn{4}{c}{Graphic} \\
        \cmidrule(lr){2-3} \cmidrule(lr){4-7} \textbf{Temperature} & Occ $\downarrow$            & Rea $\downarrow$           & Align $\downarrow$ & Und $\uparrow$ & Ove $\downarrow$ & FID$\downarrow$ \\
        \midrule
        0                                        & 0.1956                      & 0.0210                     & \textbf{0.0005}             & 0.76  & 0.0032           & 2.52            \\
        0.7                                        & 0.1482                      & 0.0162                     & 0.0010             & \textbf{0.97}  & 0.0010           & \textbf{2.00}            \\
        \rowcolor{RoyalBlue!10}1.4 (Ours)                                        & \textbf{0.1304}                      & 0.0134                     & 0.0017             & \textbf{0.97}  & 0.0012           & 2.13            \\
        2.0                                        & 0.1319                      & \textbf{0.0131}                     & 0.0027             & 0.96  & \textbf{0.0009}           & 2.12            \\
        \bottomrule
    \end{tabular}%
    }
    \caption{Comparison of results across different temperature settings of LVLM on the PKU test split.}
    \label{table:temperature}
\end{table}

\begin{figure*}[t]
    \centering
    \includegraphics[width=1\linewidth]{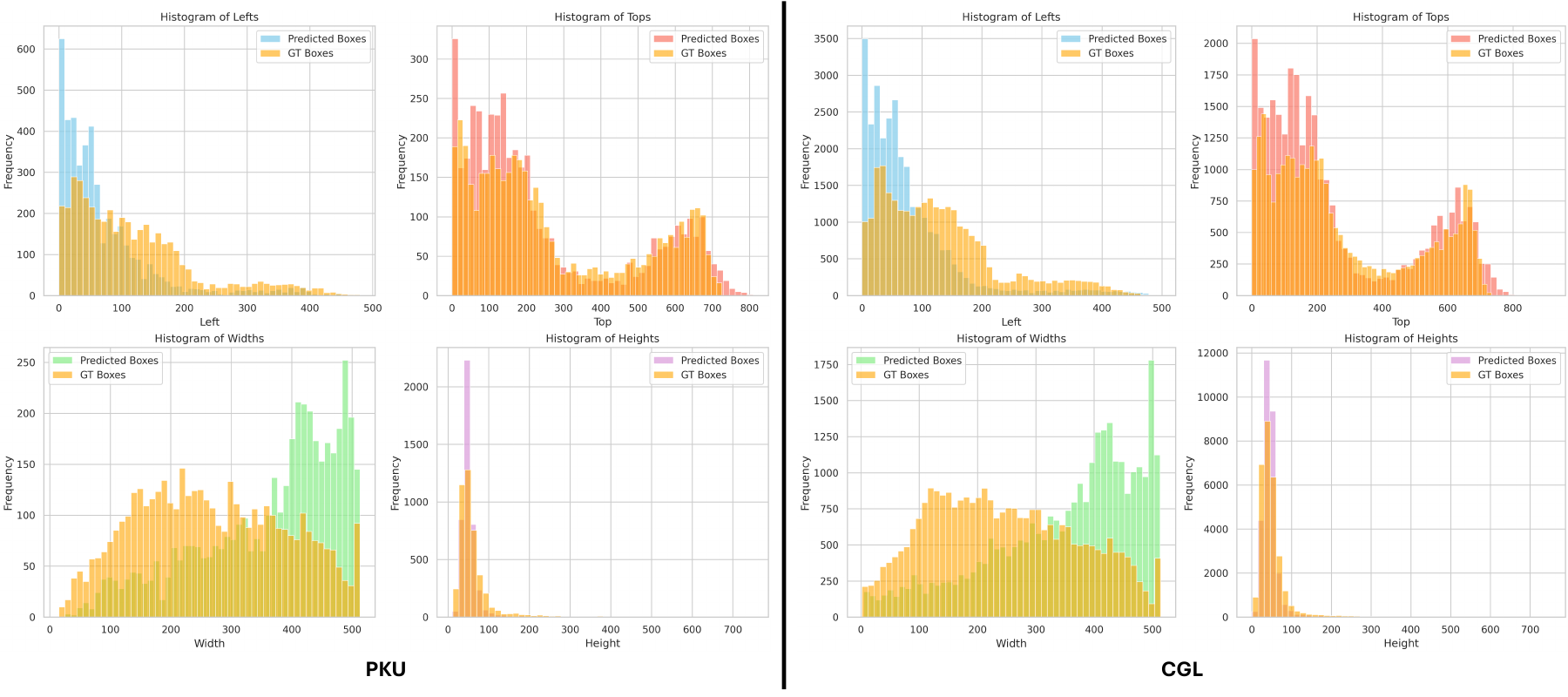}
    \caption{The distribution of generated layout results (Predicted Boxes) (coordinates: left, top, width, and height) and ground truth (GT Boxes) for the unconstrained layout generation task on the PKU and CGL test splits.}
    \label{fig:distribution}
\end{figure*}

\section{VASCAR vs. Constraint Optimization.} 
We further question whether correction similar to VASCAR is achievable via constraint-optimization
techniques \cite{kikuchi2021constrained, chentowards} that explicitly minimize the layout metrics, such as Align. We compare VASCAR and the post-processing introduced in LACE \cite{chentowards} applied to our initial layouts (\ie, by the initial prompt). LACE considers Align and Ove, which are differentiable, as the loss and minimize them with gradient decent. Obviously, as seen in \cref{table:lace}, LACE's post-processing failed in improving the non-differentiable metrics. VASCAR can be seen as a gradient-free
variant of such post-processing, and it has a clear advantage of reducing non-differential metrics, avoiding severe degradation thanks to LVLM's tendency to generate plausible layouts.

\section{Further Analyses of VASCAR}
In this section, we present additional experiments on VASCAR (Gemini), using a fixed setting of $I = 5$ and the PKU test split.

\subsection{Impact of ICL Example Retriever}
We evaluated different approaches to ICL retrieval. \textit{Random} means selecting ICL examples randomly from $\mathcal{D}$. \textit{DreamSim} is to use DreamSim \cite{fu2024dreamsim} retriever in align with Horita \etal \cite{horita2024retrieval}. \textit{IoU} refers to use IoU of saliency map. The results are shown in \cref{table:icl_selection}, indicating that \textit{Random} selection achieved the worst performance, demonstrating
that the choice of ICL examples is critical and should not be overlooked. Interestingly, retrieving ICL examples with the human-like \textit{DreamSim} approach also led to inferior performance compared to the \textit{IoU} selection. We argue that, in a training-free framework, selecting ICL examples based on IoU can implicitly enforce spatial information and compensate for the lack of \textit{content constraint}
within the textual prompt.

\subsection{Impact of Temperature on LVLM}
Temperature controls the level of randomness in token selection during generation. Lower temperatures are suited for prompts requiring deterministic or structured responses, while higher temperatures encourage more diverse and creative outputs \cite{gemini}. To investigate its impact on VASCAR, we experimented with different temperature settings, with results presented in \cref{table:temperature}. Our findings show that when the temperature is set to 0, the LVLM tends to generate fixed outputs, lacking diversity and resulting in significant degradation across almost all metrics except for Align. At a temperature of 0.7, the results are also suboptimal. Conversely, increasing the temperature to an upper bound of 2.0 leads to a substantial decline in Occ and Align metrics. These observations highlight that selecting an appropriate temperature is crucial for achieving effective and creative layout generation with VASCAR.

\section{Distribution of Generated Layouts}
We analyse the layouts generated by VASCAR ($I=15$) for the unconstrained task on the PKU and CGL test splits. The coordinates of the generated layouts, including left, top, width, and height, are restored to the original image size (513 $\times$ 750), as shown in \cref{fig:distribution}. The results indicate that the generated layouts are predominantly positioned in the upper-left region (0–100 pixels) with a large width (typically 400–500 pixels). From the figures showing the distribution of the top coordinates, there is sparse distribution in the middle regions of the canvas, suggesting that VASCAR tends to avoid placing layouts in these areas, which are often occupied by the main content.

\section{More Visual Examples}
\begin{figure*}[t]
    \centering
    \includegraphics[width=1\linewidth]{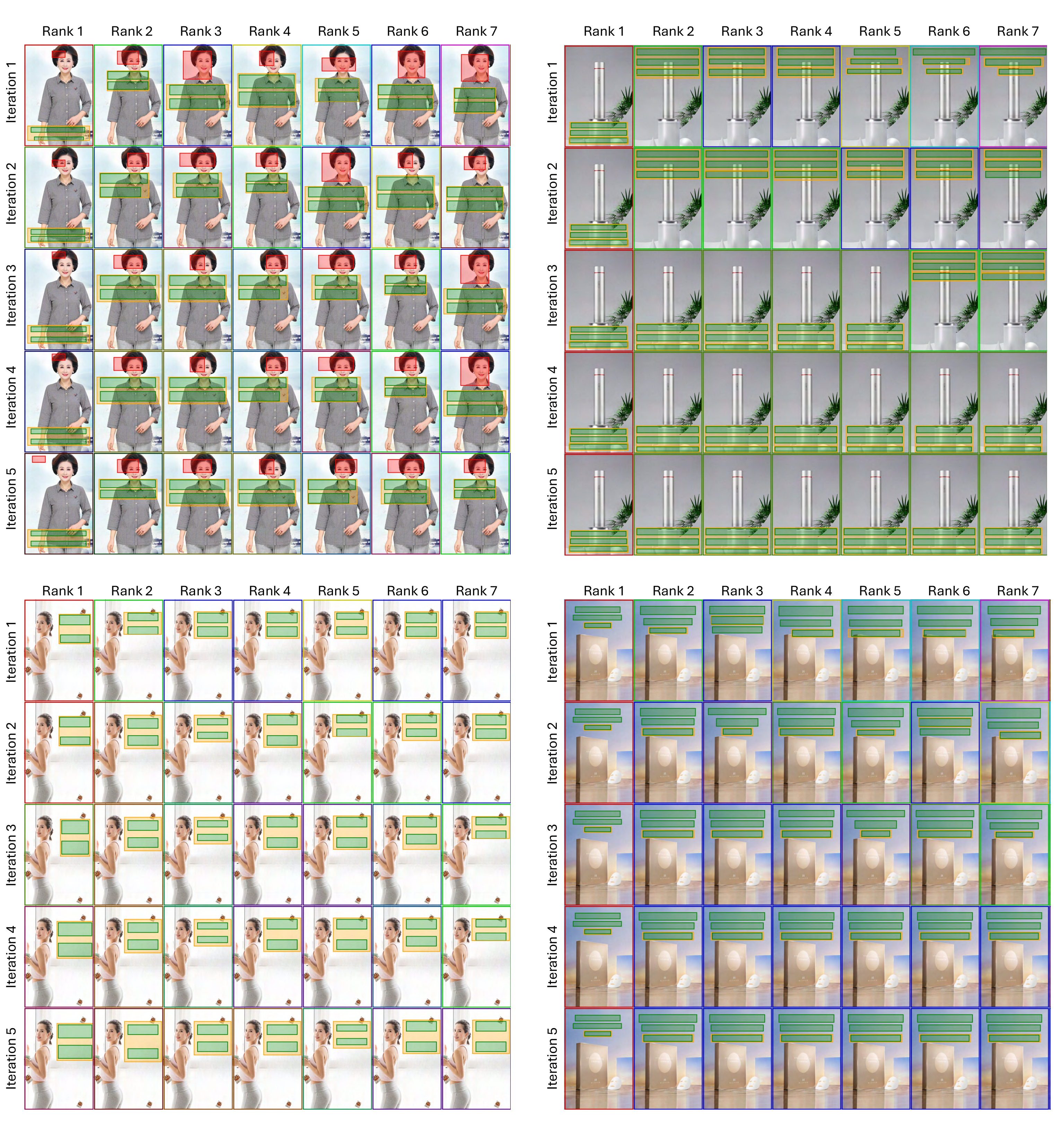}
    \caption{Visual examples of self-correction of VASCAR (GPT-4o, $I=5$).}
    \label{fig:corrections_sup}
\end{figure*}

\begin{figure*}[t]
    \centering
    \includegraphics[width=0.95\linewidth]{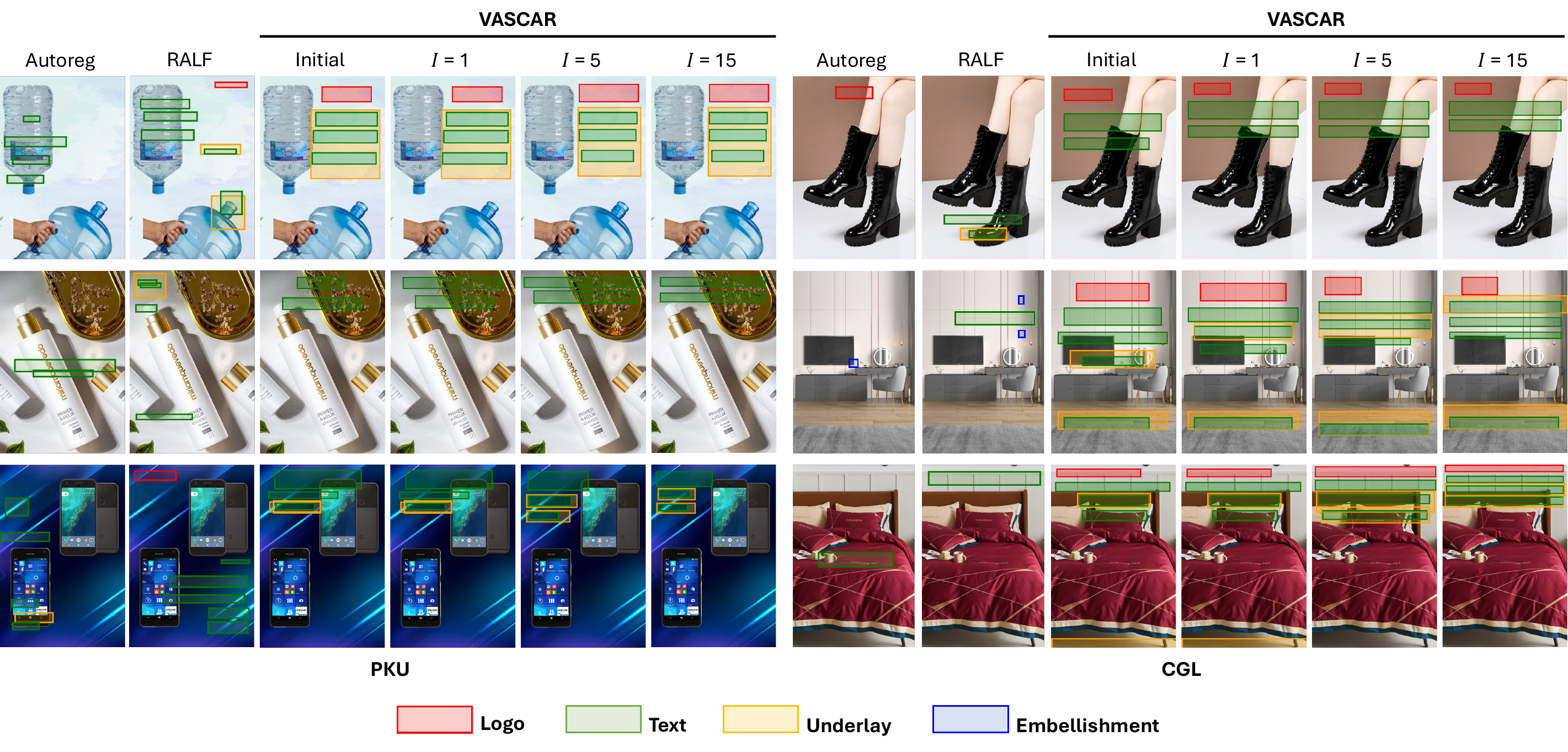}
    \caption{Visual examples of unannotated test split of PKU and CGL datasets.}
    \label{fig:unannotated_examples}
\end{figure*}

\begin{figure*}[t]
    \centering
    \includegraphics[width=0.95\linewidth]{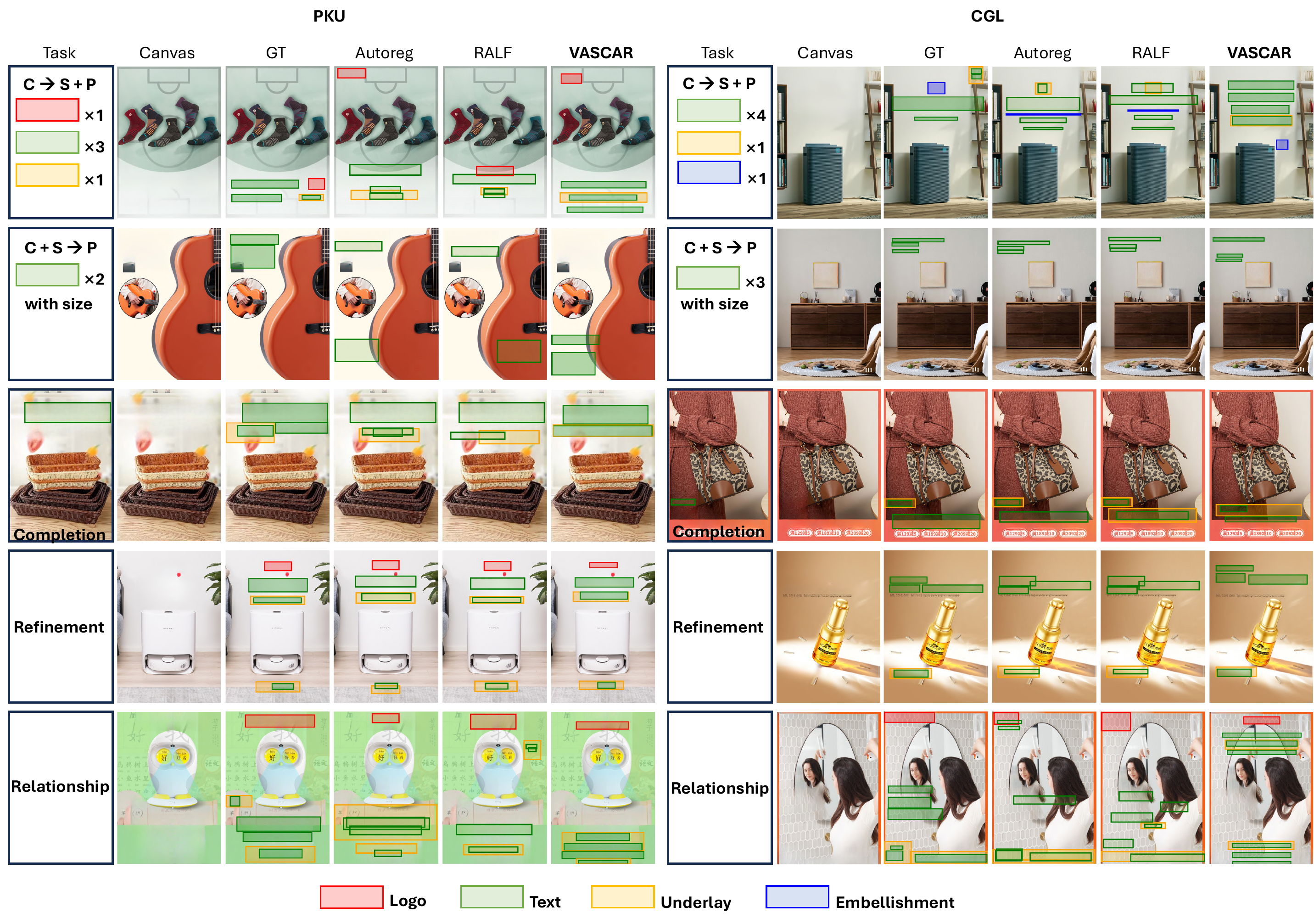}
    \caption{The visual examples for the constrained layout generation tasks on the PKU and CGL test splits. The arrangements include the canvas, ground truth (GT), Autoreg \cite{horita2024retrieval}, RALF \cite{horita2024retrieval}, and VASCAR ($I=15$).}
    \label{fig:constrained_visual_examples}
\end{figure*}

We provide additional visual examples of self-correction performed using GPT-4o in \cref{fig:corrections_sup}. Additionally, \cref{fig:unannotated_examples} presents more visual examples from the unannotated test splits of the PKU and CGL datasets. Moreover, \cref{fig:constrained_visual_examples} highlights visual examples from various constrained layout generation tasks on the PKU and CGL test splits.

\section{Prompt Examples}
\begin{figure*}[t]
    \centering
    \includegraphics[width=0.7\linewidth]{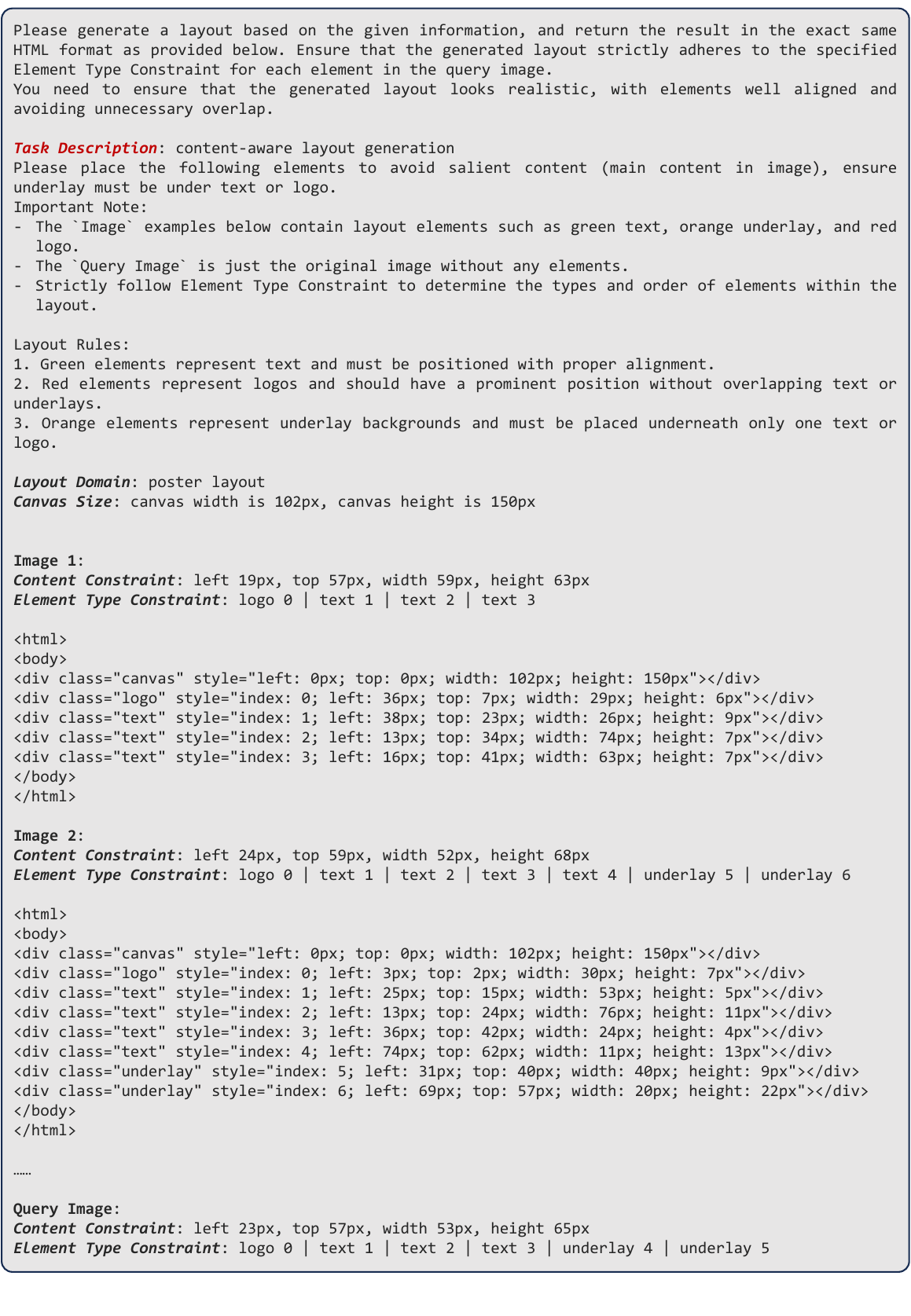}
    \caption{A prompt example of \textbf{C $\rightarrow$ S + P} task.}
    \label{table:GenC_prompt}
\end{figure*}

\begin{figure*}[t]
    \centering
    \includegraphics[width=0.7\linewidth]{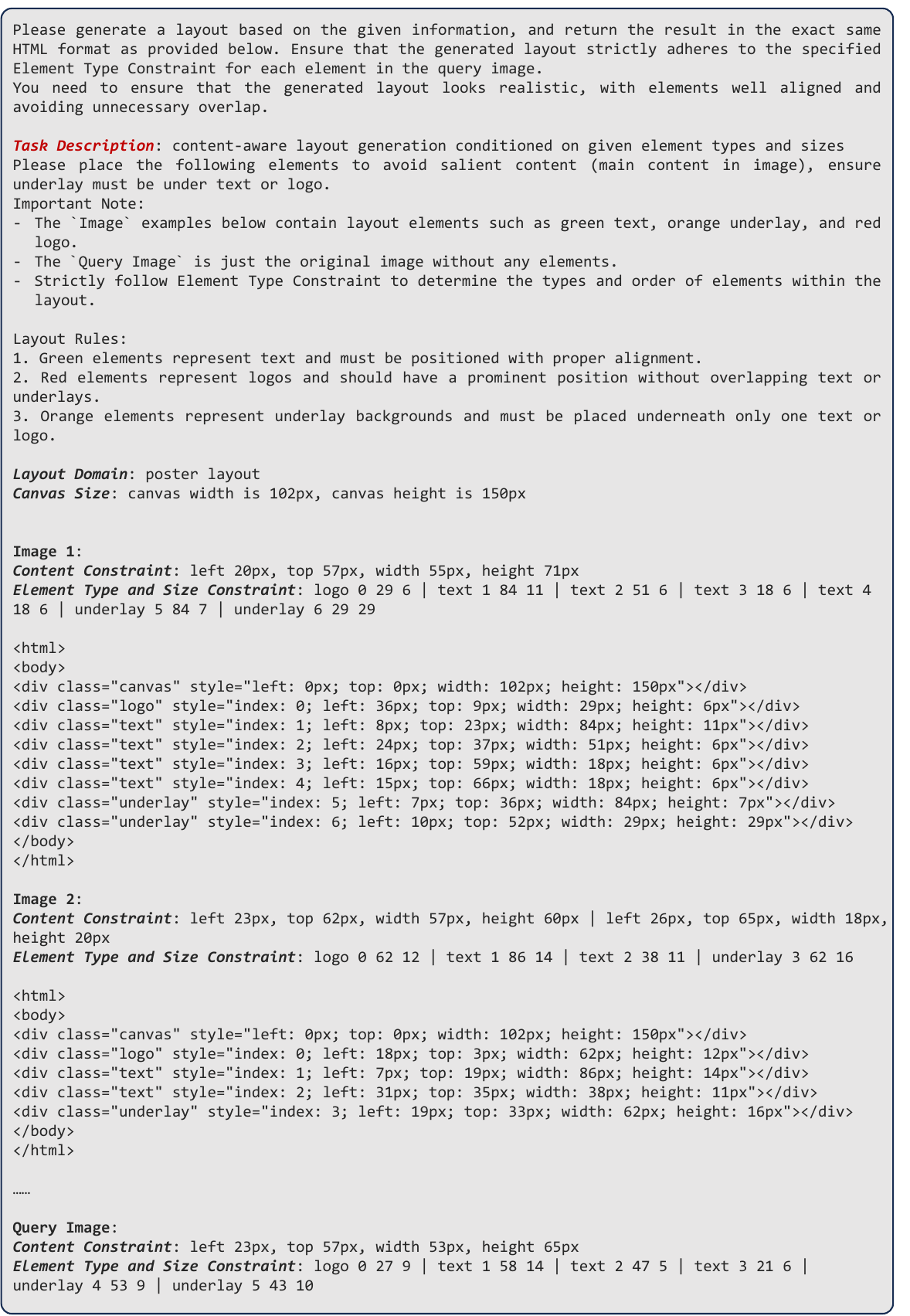}
    \caption{A prompt example of \textbf{C + S$\rightarrow$P} task.}
    \label{table:GenCS_prompt}
\end{figure*}

\begin{figure*}[t]
    \centering
    \includegraphics[width=0.7\linewidth]{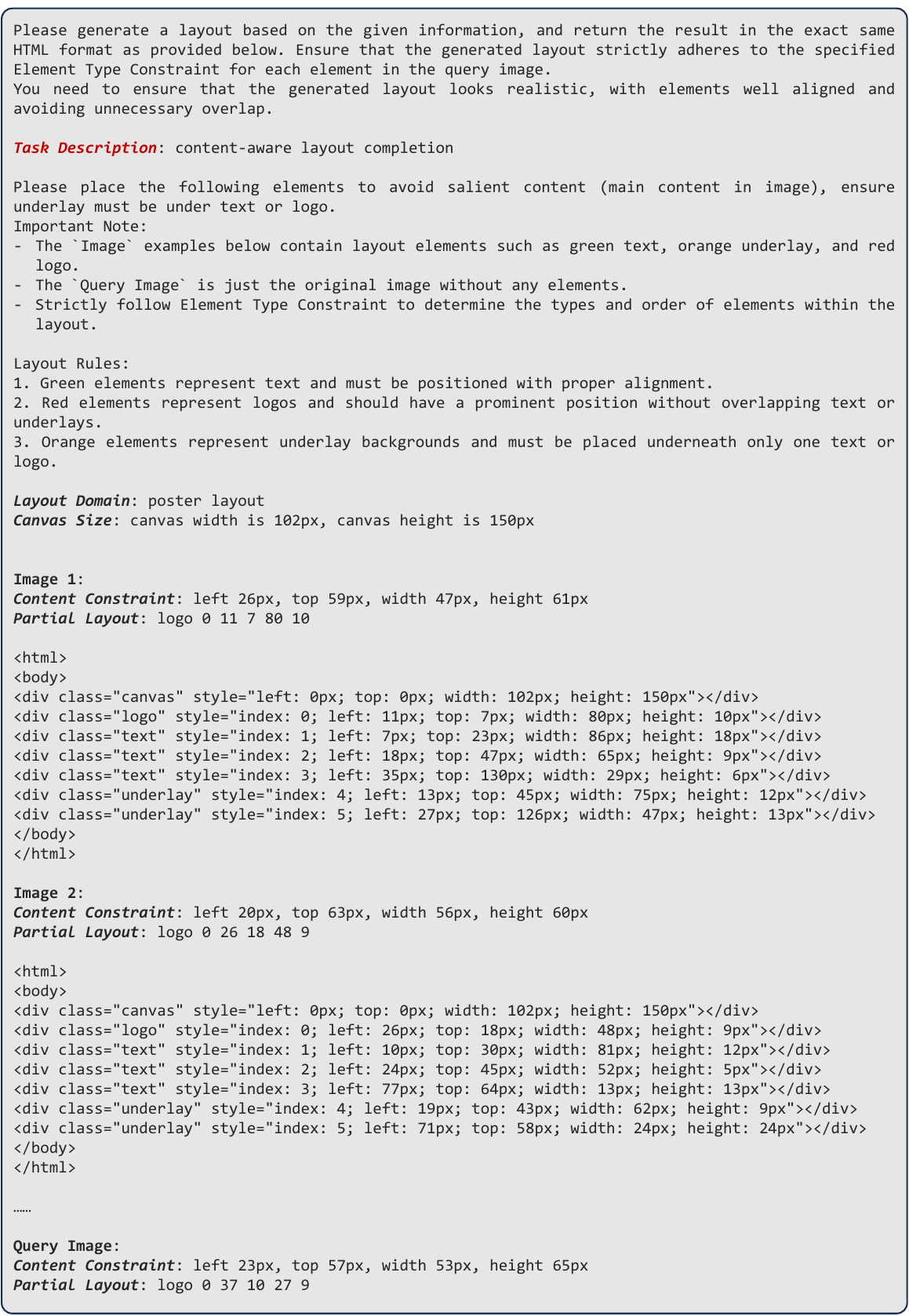}
    \caption{A prompt example of \textbf{Completion} task.}
    \label{table:completion_prompt}
\end{figure*}

\begin{figure*}[t]
    \centering
    \includegraphics[width=0.7\linewidth]{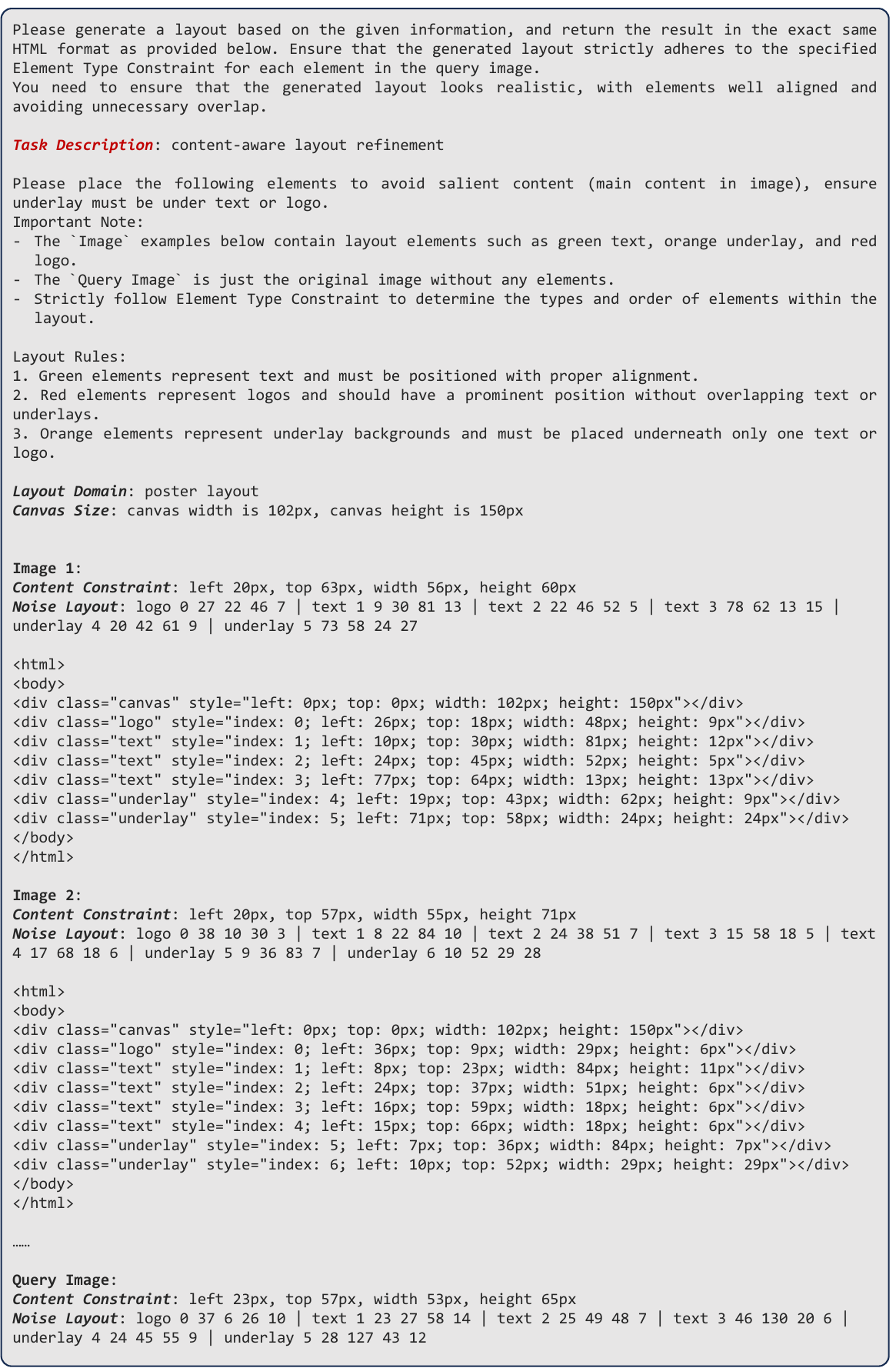}
    \caption{A prompt example of \textbf{Refinement} task.}
    \label{table:refine_prompt}
\end{figure*}

\begin{figure*}[t]
    \centering
    \includegraphics[width=0.65\linewidth]{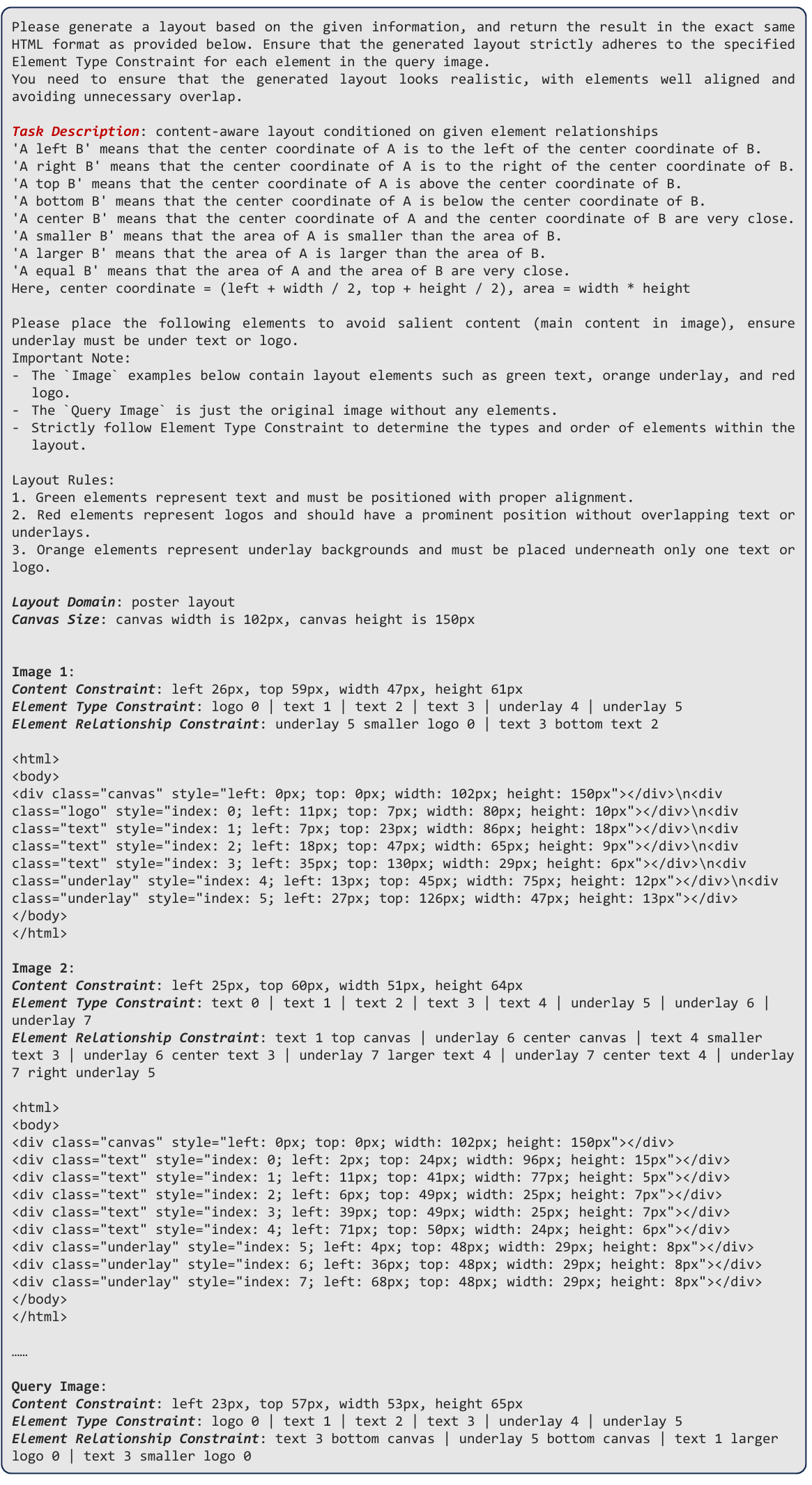}
    \caption{A prompt example of \textbf{Relationship} task.}
    \label{table:GenR_prompt}
\end{figure*}

\begin{figure*}[t]
    \centering
    \includegraphics[width=0.7\linewidth]{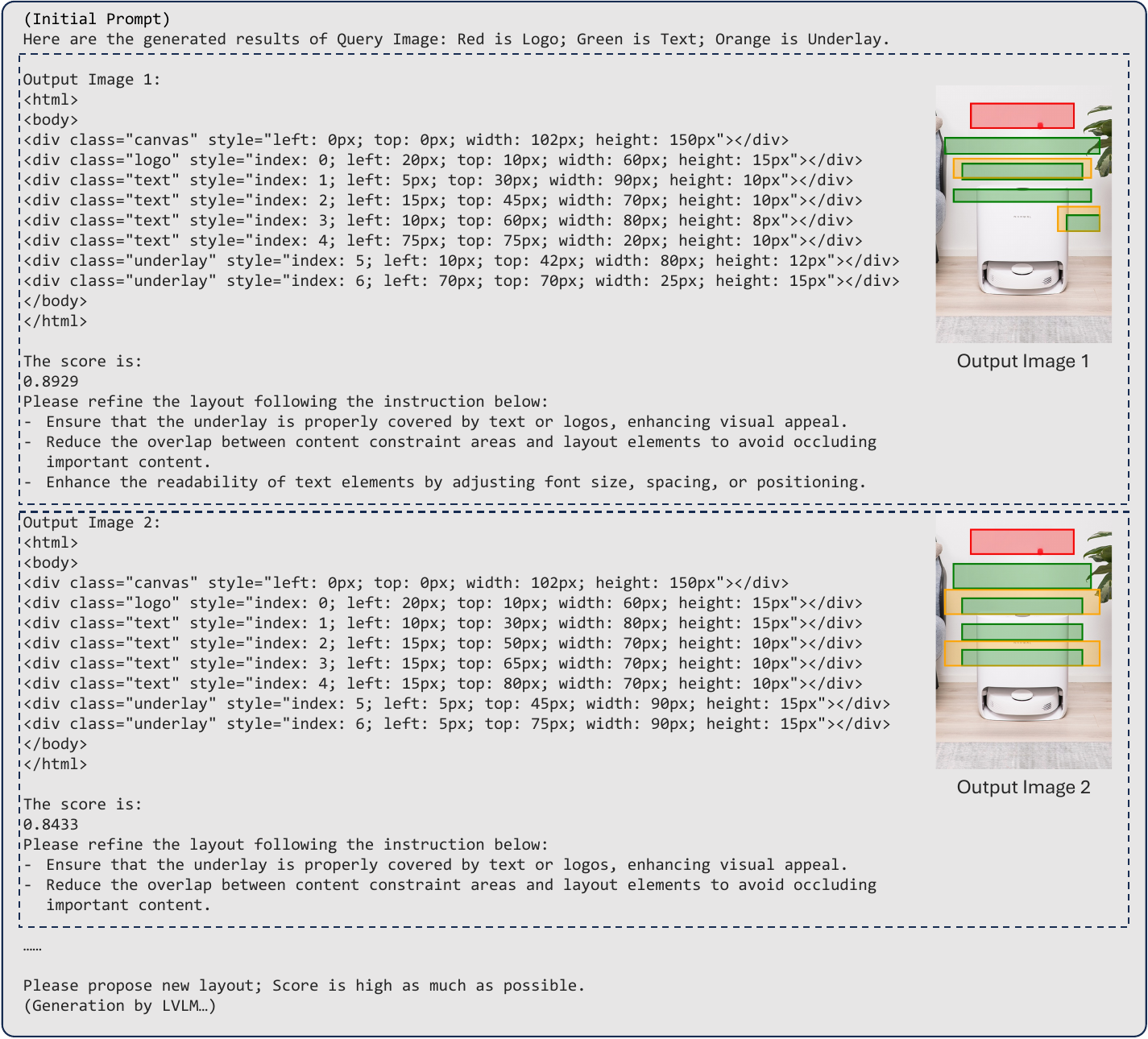}
    \caption{A prompt example of optimized prompt.}
    \label{table:optimized_prompt}
\end{figure*}

For various constrained layout generation tasks, we adopt different prompts as outlined in \cite{lin2024layoutprompter}. Here, we provide detailed examples of these prompts (excluding images for simplicity) for the following tasks: \textbf{C $\rightarrow$ S + P} (\cref{table:GenC_prompt}), \textbf{C + S $\rightarrow$ P} (\cref{table:GenCS_prompt}), \textbf{Completion} (\cref{table:completion_prompt}), \textbf{Refinement} (\cref{table:refine_prompt}), and \textbf{Relationship} (\cref{table:GenR_prompt}). Additionally, we present an optimized prompt example in \cref{table:optimized_prompt}.

\end{document}